\documentclass{article}

\PassOptionsToPackage{numbers, compress}{natbib}

\usepackage[preprint]{neurips_2026}

\usepackage[utf8]{inputenc} 
\usepackage[T1]{fontenc}    
\usepackage{hyperref}       
\usepackage{url}            
\usepackage{booktabs}       
\usepackage{graphicx}       
\usepackage{multirow}       
\usepackage{array}          
\usepackage{amsmath,amssymb} 
\usepackage{amsfonts}       
\usepackage{nicefrac}       
\usepackage{microtype}      
\usepackage{xcolor}         
\usepackage{caption}        

\usepackage{hyperref}
\usepackage[dvipsnames,table]{xcolor}
\usepackage{xspace}
\usepackage{cleveref}
\usepackage{tikz}

\newcommand{\boldstart}[1]{\vspace{0pt}\noindent\textbf{#1}}

\newcommand*{\ourmethod}{{\texttt{InvSplat}}\@\xspace}

\newcommand{\best}[1]{\textbf{#1}}
\newcommand{\second}[1]{\underline{#1}}
\usepackage{wrapfig}

\title{InvSplat: Inverse Feed-Forward Scene Splatting}

\usepackage{hyperref}
\hypersetup{
    colorlinks=true,
    urlcolor=blue,     
    linkcolor=blue,    
    citecolor=blue     
}

\usepackage{authblk}

\author[1]{Polina Karpikova}
\author[1]{Wenjing Bian}
\author[1,2]{Haofei Xu}
\author[1]{Hendrik Lensch}
\author[1]{Andreas Geiger}
\affil[ ]{$^1$University of Tübingen, Tübingen AI Center \quad $^2$ETH Zurich}

\begin{document}

\maketitle

\vspace{-2em}
\begin{figure}[h]
    \centering
    \includegraphics[width=\linewidth]{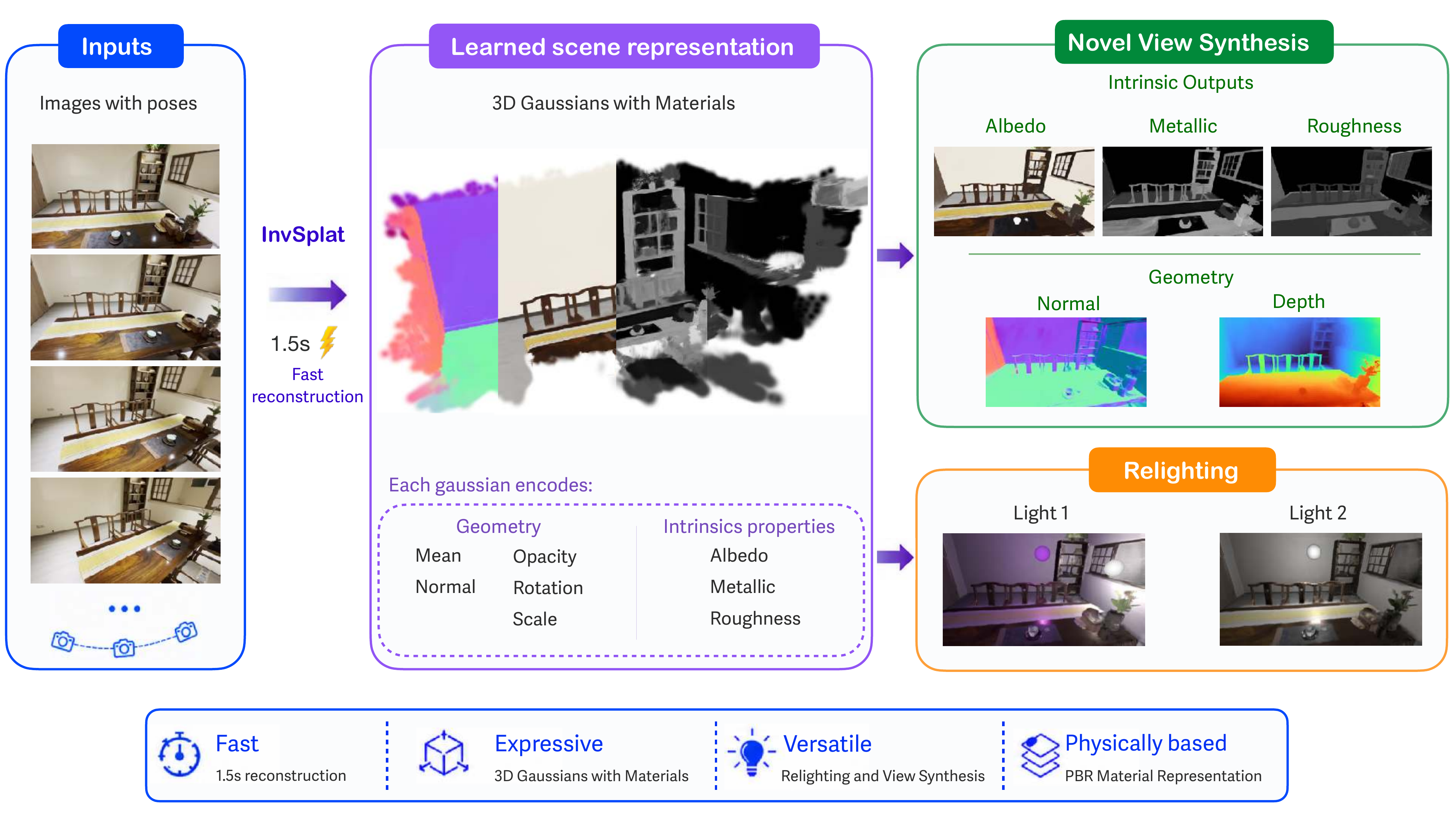}
    \vspace{-1.5em}
    \caption{\ourmethod Overview. Given a set of posed images, \ourmethod reconstructs both the 3D scene geometry and material parameters in real time, enabling novel view synthesis and relighting.}
    \label{fig:teaser}
\end{figure}
\vspace{-1.5em}

\begin{abstract}
Inverse rendering aims to recover both 3D geometry and physically meaningful material properties from images, enabling applications such as relighting and novel view synthesis. 
Optimization-based methods achieve high fidelity but require costly per-scene fitting, while image-space learning-based approaches often suffer from multi-view inconsistencies and lack an explicit 3D representation for stable novel view rendering.
We present a feed-forward multi-view reconstruction framework for inverse rendering that directly predicts a structured 3D Gaussian representation with intrinsic material attributes. 
Each Gaussian primitive is parameterized by mean, normal, opacity, rotation, scale, \emph{albedo}, \emph{metallic}, and \emph{roughness}, enabling a disentangled and physically grounded scene representation. 
Our model integrates priors from a material estimation network with a multi-view 3D reconstruction backbone, allowing joint prediction of geometry and reflectance parameters in a single forward pass.
Experiments on synthetic and real-world datasets demonstrate improved multi-view consistency compared to 2D baselines, accurate material recovery, and stable novel view rendering. 
Our representation further supports physically-based relighting and more faithful modeling of view-dependent effects compared to existing RGB-based feed-forward reconstruction methods. Our project webpage is: \url{https://poliik.github.io/invsplat/}.

\end{abstract}
\section{Introduction}


Inverse rendering aims to recover 3D geometry and physically meaningful surface reflectance from images. It forms a foundation for physically based rendering pipelines, enabling downstream applications such as relighting, material editing, and realistic AR/VR content insertion. 

Despite substantial recent progress, existing inverse rendering methods continue to face challenges in simultaneously achieving high efficiency, physical interpretability, and multi-view consistency.
Optimization-based approaches~\cite{liang2024gs,iris,zhang2021nerfactor,azinovic2019inverse,boss2021nerd} can yield accurate decompositions of geometry and reflectance, but they typically rely on computationally expensive per-scene optimization, limiting their practical value for real-world applications.
Learning-based methods~\cite{li2018learning,li2020inverse,zhu2022learning,mvinverse} estimate material properties from single- or multi-view images using feed-forward networks or iterative diffusion-based denoising~\cite{diffusionrenderer,dnf_intrinsics}. While these approaches significantly improve runtime efficiency compared to optimization-based pipelines, existing methods operate predominantly in image space (e.g., predicting per-view intrinsic maps) and lack an explicit 3D scene representation. Consequently, they struggle to support robust novel view synthesis and are prone to producing view-inconsistent geometry and material predictions (\Cref{fig:consistency_main}), which hinders physically grounded editing and relighting.

\noindent
\begin{minipage}[t]{0.56\textwidth}
\vspace{-2pt}

In parallel, recent feed-forward 3D reconstruction methods~\cite{worldmirror, depthanything3, wang2025vggt, Wang_2024_CVPR, xu2025resplat} have demonstrated that multi-view images can be converted into explicit 3D scene representations in a single forward pass. However, these approaches primarily predict geometry together with RGB appearance, rather than intrinsic, physically meaningful material parameters. While some approaches~\cite{xu2025depthsplat, worldmirror} can model low- to mid-frequency view-dependent effects, they often struggle to reproduce sharp specular highlights. Moreover, illumination is typically implicitly baked into the predicted appearance, preventing explicit relighting or material changes.

\end{minipage}\hfill
\begin{minipage}[t]{0.42\textwidth}
\vspace{-2pt}
\centering
\setlength{\tabcolsep}{2pt}
\resizebox{\linewidth}{!}{%
\begin{tabular}{@{}lccc@{}}
\toprule
\textbf{Method} & \textbf{Paradigm} & \textbf{Consistent} & \textbf{NVS} \\
\midrule
IRIS~\cite{iris} & optimization & \checkmark & \checkmark \\
Intrinsic Image Fusion~\cite{intrinsicimagefusion} & optimization & \checkmark & \checkmark \\
DiffusionRenderer~\cite{diffusionrenderer} & diffusion & $\sim$ & $\times$ \\
DNF-Intrinsic~\cite{dnf_intrinsics} & diffusion/flow & $\sim$ & $\times$ \\
MVInverse~\cite{mvinverse} & feed-forward & $\sim$ & $\times$ \\
\midrule
InvSplat (Ours) & feed-forward & \checkmark & \checkmark \\
\bottomrule
\end{tabular}%
}
\captionof{table}{High-level methods comparison. ``Consistent'' denotes multi-view consistency; $\sim$ indicates partial/limited support; $\times$ indicates not supported by design.}
\label{tab:method_summary}
\end{minipage}

To address these limitations, we introduce a feed-forward inverse rendering model that directly predicts a physically based 3D Gaussian scene representation from posed multi-view images. Thanks to our 3D Gaussian representation, our model is by design multi-view consistent and can naturally support real-time rendering. In addition, we predict all Gaussian parameters together with intrinsic material attributes in a single forward pass, removing the need for the expensive per-scene optimization or iterative multi-step diffusion used in previous work. This makes our model highly efficient. In \Cref{tab:method_summary}, we provide a conceptual comparison with representative prior methods, showing that our model is the first to simultaneously achieve efficient feed-forward inference, high multi-view consistency, and high-quality novel view synthesis.

More specifically, our method outputs Gaussian primitives parameterized by geometry and opacity, along with intrinsic material attributes (\emph{albedo}, \emph{metallic}, and \emph{roughness}). The reconstructed scene can be rendered from arbitrary camera poses using a differentiable Gaussian rasterizer, enabling consistent inverse rendering and real-time novel view synthesis within a unified framework.
In contrast to image-based inverse rendering approaches that estimate intrinsic properties only for the observed views, our method recovers an explicit 3D representation that supports rendering from arbitrary viewpoints. Since a primary motivation for disentangling illumination from material properties is to enable relighting, we demonstrate this capability with a small point-light renderer for generating relit results.


We summarize our contributions as follows: 
\begin{itemize} \setlength{\itemsep}{0pt} \setlength{\parskip}{0pt} 
 \item We introduce \ourmethod, the first feed-forward framework for scenes that predicts physically based 3D Gaussian primitives with intrinsic material parameters from multi-view images in a single forward pass. \item \ourmethod integrates priors from image-based material estimation and 3D reconstruction models into a unified network, producing high-quality 3D material properties with strong generalization to real-world data. 
\item We achieve novel view synthesis in a reconstructed 3D Gaussian splatting scene with decomposed materials, thus enabling relighting applications. 
\end{itemize}
\section{Related Work}


\textbf{Optimization/Learning-based approaches.}
Optimization-based inverse rendering fits a scene representation to each capture via iterative updates.
It can be accurate and physically grounded, but inference is slow and must be repeated for every new scene.
For example, IRIS~\cite{iris} estimates geometry, physically based materials, spatially varying HDR lighting, and camera response from posed multi-view LDR images via an iterative optimization pipeline.
Intrinsic Image Fusion~\cite{intrinsicimagefusion} is another optimization-based multi-view approach that fuses per-view intrinsic priors into a consistent, low-dimensional material space and then refines it with inverse path tracing. Recently, MVInverse~\cite{mvinverse} scaled up material prediction by training a large transformer model from multi-view inputs.

\textbf{Diffusion-based approaches.}
Diffusion models provide strong image priors for inverse problems and have been used to predict intrinsic properties or G-buffers. DiffusionRenderer~\cite{diffusionrenderer} leverages video diffusion priors to estimate G-buffers from real videos and additionally trains a neural renderer from G-buffers to enable editing workflows; however, it inherits the computational cost and stability challenges of diffusion sampling. DNF-Intrinsic~\cite{dnf_intrinsics} proposes a deterministic alternative (flow-matching / noise-free diffusion) together with a generative rendering constraint to improve faithfulness, but it remains substantially heavier than a single forward pass. Moreover, without explicit 3D coupling, such image-space diffusion approaches can produce predictions that drift across viewpoints.

\paragraph{Feed-forward Scene Reconstruction.}

Recent advancements in 3D scene representation have seen a paradigm shift towards feed-forward 3D Gaussian Splatting (3DGS) architectures, which bypass time-consuming per-scene optimization~\cite{kerbl20233d} to generate high-fidelity novel views. pixelSplat~\cite{charatan2024pixelsplat} pioneered this direction by learning to reconstruct radiance fields from sparse image pairs using dense probability distributions. Subsequent works rapidly improved both efficiency and geometric accuracy; MVSplat~\cite{chen2024mvsplat} introduced plane-sweeping cost volumes for fast multi-view feature matching, while DepthSplat~\cite{xu2025depthsplat} integrated pre-trained depth features to inject stronger geometric priors. To address the limitations of single-pass inference, ReSplat~\cite{xu2025resplat} proposed a recurrent framework that iteratively refines 3D Gaussian primitives using rendering error feedback. More recently, researchers have expanded input constraints and multi-task capabilities, with NoPoSplat~\cite{ye2024no} directly predicting scene Gaussians from unposed images, and WorldMirror~\cite{worldmirror} introducing a universal framework capable of simultaneously outputting multiple 3D representations in a single forward pass.

While these feed-forward methods have achieved remarkable success in RGB novel view synthesis, they are fundamentally designed for forward rendering tasks. They focus solely on synthesizing novel appearances and do not disentangle the scene into its underlying intrinsic properties, such as materials, lighting, and precise geometric normals. In contrast, our work addresses the largely underexplored domain of inverse feed-forward modeling. We propose \ourmethod, the first feed-forward Gaussian splatting framework specifically formulated for inverse scene reconstruction. By extending the efficiency of feed-forward architectures into the realm of inverse rendering, our method uniquely enables instantaneous scene decomposition without the need for expensive per-scene optimization.

\begin{figure*}[t]
  \centering
\includegraphics[width=\textwidth]{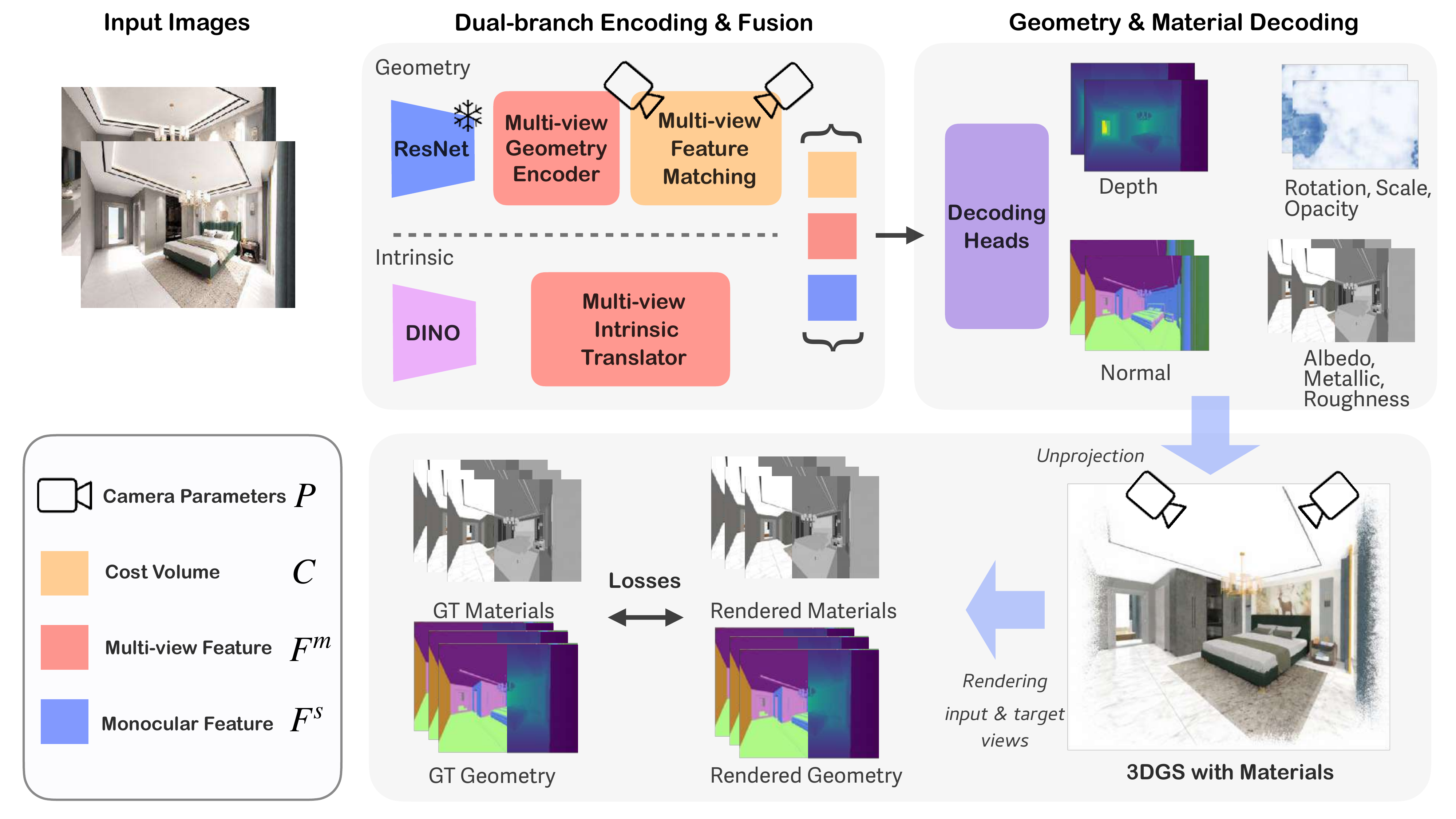}
  \caption{\textbf{Method Overview}. Our feed-forward multi-view model predicts a physically based 3D Gaussian scene representation (geometry + material parameters) and enforces cross-view consistency through differentiable rendering. }
  \label{fig:method}
\end{figure*}
\section{Method}

Given $N$ input images $\{I_i\}_{i=1}^N$ capturing a shared scene and their corresponding camera parameters $\{\mathbf{P}_i\}_{i=1}^N$ (including intrinsics and extrinsics), our goal is to jointly recover the scene geometry $\mathcal{G}$ and intrinsic material properties $\mathcal{M}$. 
We employ a feed-forward network that predicts all scene parameters in a single forward pass:
\begin{equation}
(\mathcal{G}, \mathcal{M}) 
= f_{\theta}\big(\{I_i\}_{i=1}^{N}, \{\mathbf{P}_i\}_{i=1}^N\big),
\end{equation}
where $f_{\theta}$ is parameterized by $\theta$. We first introduce the scene representation in \Cref{subsec:scene_representation_intrinsic}, followed by the feed-forward inverse rendering framework in \Cref{subsec:inverse_rendering}, and finally the training objectives in \Cref{subsec:training}.

\subsection{Scene Representation with Intrinsic Properties}
\label{subsec:scene_representation_intrinsic}

We adopt 3D Gaussian Splatting (3DGS)~\cite{kerbl20233d} to represent the scene using $M$ Gaussian primitives. 
Each Gaussian $j$ is parameterized by its mean $\boldsymbol{\mu}_j \in \mathbb{R}^3$, rotation quaternion $\mathbf{q}_j \in \mathbb{R}^4$, scale $\mathbf{s}_j \in \mathbb{R}^3$, and opacity $\sigma_j \in [0,1]$. While the original formulation models appearance using spherical harmonics coefficients, we instead focus on intrinsic material properties. Specifically, each Gaussian is associated with diffuse albedo $\mathbf{a}_j \in [0,1]^3$, metallicity $m_j \in [0,1]$, and roughness $r_j \in [0,1]$. We further augment each Gaussian with a surface normal $\mathbf{n}_j \in \mathbb{R}^3$, which enables high-quality physically based shading and relighting. The scene is thus represented as
\[
\mathcal{G} = \{ (\boldsymbol{\mu}_j, \mathbf{q}_j, \mathbf{s}_j, \sigma_j, \mathbf{n}_j) \}_{j=1}^{M},
\qquad
\mathcal{M} = \{ (\mathbf{a}_j, m_j, r_j) \}_{j=1}^{M}.
\]

\subsection{Feed-forward Inverse Rendering Architecture}
\label{subsec:inverse_rendering}

We propose a feed-forward inverse rendering framework that jointly predicts scene geometry as a 3D Gaussian representation with intrinsic material attributes from multi-view posed images. 
As illustrated in \Cref{fig:method}, the architecture follows a dual-branch design: a \emph{Geometry branch} that builds an explicit multi-view geometric representation from the input views, and an \emph{Intrinsic branch} that extracts cross-view intrinsic features. Features from both branches are then consumed by a set of decoding heads that together produce a unified 3D Gaussian representation with intrinsic material attributes.

\paragraph{Geometry branch.}
To recover accurate 3D structure, we adopt a stereo-inspired multi-view pipeline similar to ReSplat~\cite{xu2025resplat}. For each input view $I_i$, we first apply a ResNet~\cite{he2015deep} image backbone to produce a multi-scale feature pyramid $\{\mathbf{F}^{s}_i\}$. The deepest pyramid level is fed into the \emph{Multi-view Geometry Encoder}, a transformer that performs cross-view self-attention to aggregate information across views and produce view-aware geometry features. These features are then passed to the \emph{Multi-view Feature Matching} module, which warps them across views using the input camera poses and constructs a depth-candidate cost volume $\mathbf{C}_i$ encoding cross-view correspondences. The shallower scales of the ResNet pyramid bypass the matching step and are forwarded directly to the decoding heads, where they provide higher-resolution image cues for dense per-pixel prediction.

\paragraph{Intrinsic branch.}
In parallel, we extract high-level multi-view intrinsic features through attention, similar to previous works~\cite{mvinverse, wang2025vggt}. 
 We first use a DINOv2~\cite{oquab2023dinov2} ViT-L/14 with register features to encode each input view independently into patch features. These features are then refined by the \emph{Multi-view Intrinsic Translator}, a 36-block transformer that alternates intra-view and inter-view self-attention, exchanging information across views. Set of features $\{\mathbf{F}^m_i\}_{i=1}^N$ from uniformly spaced translator layers are used for the decoding heads. We refer to this transformer as a \emph{translator}, with the role of mapping the appearance features into a latent space consumable by downstream heads. During training, the translated features are learned through both the depth decoder of the geometry branch and the material decoders, serving as a shared block across decoder branches.

\paragraph{Decoding and Gaussian construction.}
We decode per-view geometric and material properties through 6 decoder heads.
 Depth $d_i$ is estimated from all features $\{\mathbf{C}_i, \mathbf{F}^m_i, \mathbf{F}^s_i\}$ with a DPT~\cite{Ranftl2020}  head, while to estimate the per-Gaussian rotation $\mathbf{q}_j$, scale $\mathbf{s}_j$, and opacity $\sigma_j$, we lift features into 3D using the predicted depth and refine them with a Point Transformer~\cite{zhao2021point,xu2025resplat} that captures local geometric context; the refined features are then mapped to Gaussian parameters by a lightweight regression head.
Material properties (albedo $\mathbf{a}_i$, metallicity $m_i$, roughness $r_i$) and the surface normals $\mathbf{n}_i$ are predicted by four DPT heads from $\{\mathbf{F}^m_i\}$. At the albedo head additional skip connections with $\{\mathbf{F}^s_i\}$ are applied, and we add a single $1\!\times\!1$ convolutional layer that matches the channel widths of $\{\mathbf{F}^s_i\}$ to those of the pretrained head. We unproject the predicted depth maps into 3D to obtain Gaussian centers $\{\boldsymbol{\mu}_j\}_{j=1}^M$, where each Gaussian inherits its image-space material predictions $(\mathbf{a}_j, m_j, r_j)$ from its source pixel, and the per-Gaussian normal $\mathbf{n}_j$ is obtained by rotating the predicted camera-space normal to world space using the input extrinsics.

\paragraph{Rendering.}
The predicted Gaussians are projected into each view using camera parameters $\{\mathbf{P}_i\}$. 
We employ a differentiable Gaussian splatting renderer to produce albedo, metallic, roughness, normal, and depth maps in a single rasterization pass.






\subsection{Training}
\label{subsec:training}
During training, for each forward path, given ${N_t}$ camera poses, we render ${N_t}$ maps for each geometric and material property, and supervise all the rendered properties with per-view ground truth.
For each view $i$, let $(\mathbf{a}_i, m_i, r_i, d_i, \mathbf{n}_i)$ denote the ground-truth albedo, metallicity, roughness, depth, and normals, and $(\hat{\mathbf{a}}_i, \hat{m}_i, \hat{r}_i, \hat{d}_i, \hat{\mathbf{n}}_i)$ be the corresponding renderings from the predicted scene. 
The overall training objective is
\begin{equation}
\mathcal{L} = 
\mathcal{L}_{\mathbf{a}} +
\mathcal{L}_{m} +
\mathcal{L}_{r} +
\mathcal{L}_{d} +
\mathcal{L}_{\mathbf{n}}.
\end{equation}

\paragraph{Material supervision.}
We supervise intrinsic components using a combination of pixel-wise and perceptual losses. For each $X \in \{\mathbf{a}, m, r\}$, we define
\begin{equation}
\mathcal{L}_{X} = \sum_{i=1}^N 
\Big( \lVert \hat{X}_i - X_i \rVert_1 
+ \lambda_{\text{lpips}}\, \mathrm{LPIPS}(\hat{X}_i, X_i) \Big).
\end{equation}

\paragraph{Affine-invariant depth loss.}
As our training dataset provides depth maps, but doesn't contain camera parameters, we supervise depth using an affine-invariant loss, inspired by MoGe \cite{wang2025moge} for self-supervised depth estimation:
\begin{equation}
\mathcal{L}_{d} = \sum_{i=1}^N \frac{1}{|\Omega_i|} \sum_{p \in \Omega_i}
\left( \delta_i(p) - \bar{\delta}_i \right)^2,
\end{equation}
where $\delta_i(p) = \log \hat{d}_i(p) - \log d_i(p)$, 
$\bar{\delta}_i = \frac{1}{|\Omega_i|} \sum_{p \in \Omega_i} \delta_i(p)$, 
and $\Omega_i$ denotes the set of valid pixels.

\paragraph{Normal loss.}
Surface normals are supervised using a cosine similarity loss:
\begin{equation}
\mathcal{L}_{n} = \sum_{i=1}^N \left( 1 - \langle \hat{\mathbf{n}_i}, \mathbf{n}_i \rangle \right).
\end{equation}





\section{Experiments}

\begin{figure*}[t]
  \centering
  \setlength{\tabcolsep}{0pt}
  \begin{tabular}{@{}*{6}{>{\centering\arraybackslash}p{0.1666\textwidth}}@{}}
    \footnotesize Input & \footnotesize  Albedo GT & \footnotesize   Ours & \footnotesize  MVInverse & \footnotesize  DiffusionRenderer & \footnotesize  DNF-intrinsic \\
  \end{tabular}\\[-2pt]
  \includegraphics[width=\textwidth]{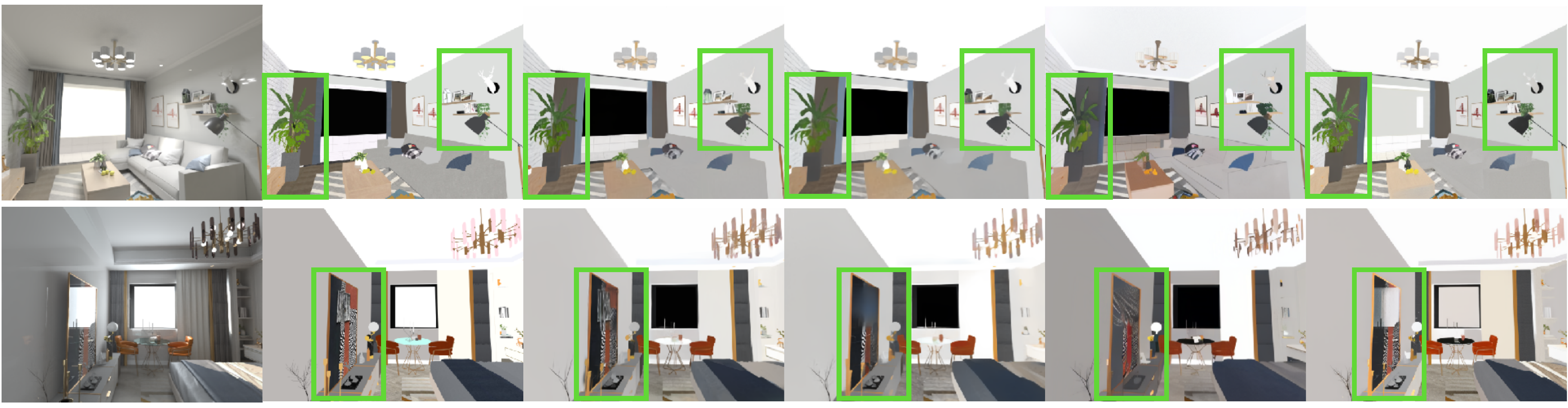}
  \caption{Qualitative reconstruction results on InteriorVerse. \ourmethod{} jointly predicts geometry and intrinsic material attributes (albedo, metallic, roughness) in a single feed-forward pass, producing multi-view consistent decompositions.
  }\vspace{-13pt}
  \label{fig:interiorverse_reconstruction}
\end{figure*}

\boldstart{Implementation details.}
We train on InteriorVerse dataset~\cite{zhu2022learning} on resolution 512x384 with curated image triplets, using two views as input views and all three views' ground truth for supervision, which includes depth, normals, and material maps. The geometric encoder is initialized from ReSplat~\cite{xu2025resplat}; we freeze the ResNet backbone during training and fine-tune the remaining components. The appearance encoder and material prediction heads are initialized from MVInverse~\cite{mvinverse}, and all associated parameters are jointly fine-tuned with the rest of the network. Training is performed for 20k steps with a batch size of 2 on a single H100 GPU, taking approximately 12 hours.

\boldstart{Baselines.}
 We compare our model against three 2D baselines: DiffusionRenderer~\cite{diffusionrenderer}, a video diffusion model; DNF-Intrinsic~\cite{dnf_intrinsics}, a single-view diffusion approach; and MVInverse~\cite{mvinverse}, a feed-forward multi-view transformer. For MVInverse, we also report a variant fine-tuned on our training data.
 Per-scene optimization baselines are not included in the comparison, since the sparse input views are insufficient for robust geometry, material and lighting disentanglement of typical existing optimization-based approaches.

\boldstart{Datasets.}
We follow the standard train/test split of InteriorVerse~\cite{zhu2022learning} and additionally evaluate on Structured3D~\cite{zheng2020structured3d}. Further details are provided in the supplementary material. We also conduct qualitative evaluations on procedurally generated scenes from Infinigen~\cite{infinigen2024indoors} and real-world images from RealEstate10K~\cite{zhou2018stereo} and DL3DV~\cite{ling2024dl3dv}. 

\boldstart{Metrics.}
Following prior work~\cite{mvinverse, dnf_intrinsics}, we assess material and normal quality on input views. To evaluate cross-view consistency, we compute a reprojection error metric: predictions are warped between to each other using ground-truth depths and ground-truth camera poses, and RMSE is computed on the resulting correspondences. 

\subsection{Evaluation}
\begin{table}[t]
\centering
\footnotesize
\setlength{\tabcolsep}{2.6pt}
\caption{
Quantitative comparison of inverse rendering performance on the synthetic InteriorVerse test set using two input views per scene. * denotes fine-tuned networks. Matching the performance of 2D pixel-aligned image networks with a unified 3D reconstruction model remains challenging. Nevertheless, our method achieves performance comparable to 2D baselines while providing an explicit 3D representation, resulting in improved cross-view consistency (~\Cref{fig:consistency_main}) and enabling novel-view rendering (~\Cref{fig:re10k_nvs}).
}
\begin{tabular}{@{}lcccccccc@{}}
\toprule
\textbf{Method} & \textbf{Type} & \multicolumn{4}{c}{\textbf{Albedo}} & \textbf{Metallic} & \textbf{Roughness} & \textbf{Normal} \\
\cmidrule(lr){3-6} \cmidrule(lr){7-7} \cmidrule(lr){8-8} \cmidrule(lr){9-9}
 & & \textbf{PSNR}$\uparrow$ & \textbf{SSIM}$\uparrow$ & \textbf{LPIPS}$\downarrow$ & \textbf{RMSE}$\downarrow$ & \textbf{RMSE}$\downarrow$ & \textbf{RMSE}$\downarrow$ & \textbf{Cosine Similarity}$\uparrow$ \\
\midrule
DiffusionRenderer~\cite{diffusionrenderer} & 2D & 17.32 & 0.800 & 0.253 & 0.1506 & 0.2971 & 0.2825 & 0.9468 \\
DNF-Intrinsic~\cite{dnf_intrinsics} & 2D & 18.64 & 0.850 & 0.211 & 0.1320 & 0.1884 & 0.2124 & 0.9261 \\
MVInverse~\cite{mvinverse} & 2D & 21.83 & 0.867 & 0.217 & 0.0887 & 0.1039 & \second{0.1252} & \best{0.9654} \\
MVInverse* & 2D & \best{22.92} & \best{0.886} & \best{0.182} & \best{0.0798} & \best{0.0985} & \best{0.1221} & \second{0.9630} \\
\midrule
\textbf{Ours} & \best{3D} & \second{22.18} & \second{0.873} & \second{0.203} & \second{0.0883} & \second{0.0993} & 0.1254 & 0.9609 \\

\bottomrule
\end{tabular}
\label{tab:interiorverse_materials}
\end{table}
\begin{figure*}[t]
  \centering
  \setlength{\tabcolsep}{0pt}
  \renewcommand{\arraystretch}{0}
  \resizebox{\textwidth}{!}{%
  \begin{tabular}{@{}cccccc@{}}
    {\footnotesize Input} & {\footnotesize Albedo NVS} & {\footnotesize Input} & {\footnotesize Albedo NVS} & {\footnotesize Input} & {\footnotesize Albedo NVS} \\
    \includegraphics[width=0.165\textwidth]{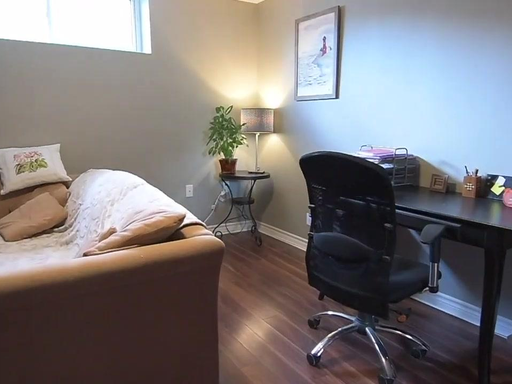} &
    \includegraphics[width=0.165\textwidth]{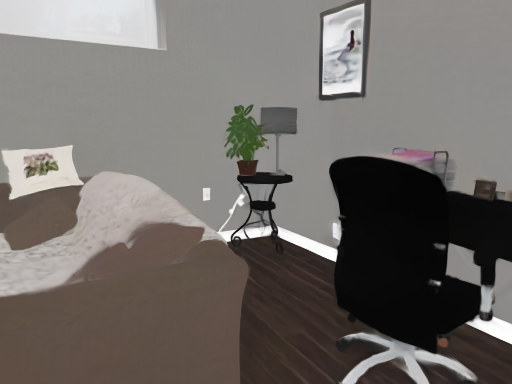} &
    \includegraphics[width=0.165\textwidth]{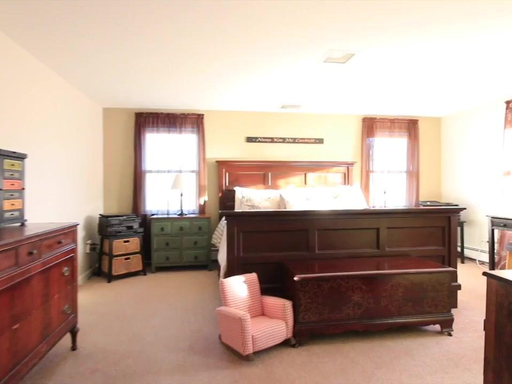} &
    \includegraphics[width=0.165\textwidth]{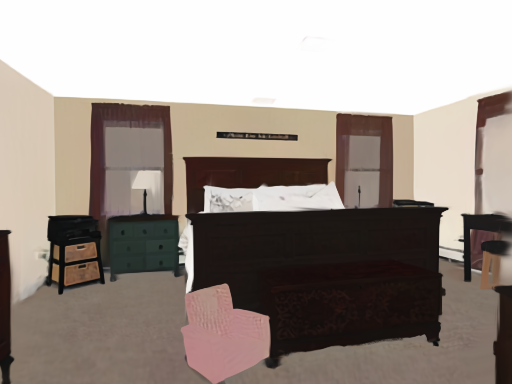} &
    \includegraphics[width=0.165\textwidth]{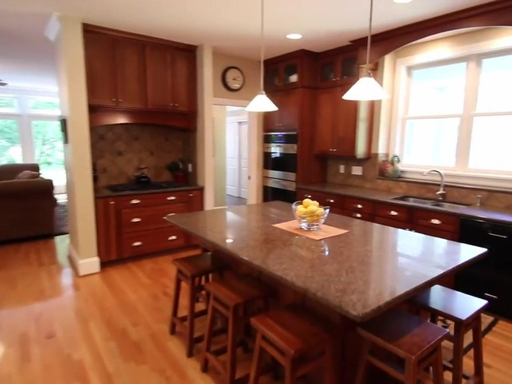} &
    \includegraphics[width=0.165\textwidth]{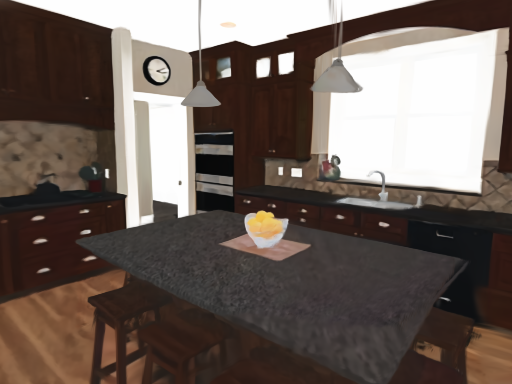} \\
    \includegraphics[width=0.165\textwidth]{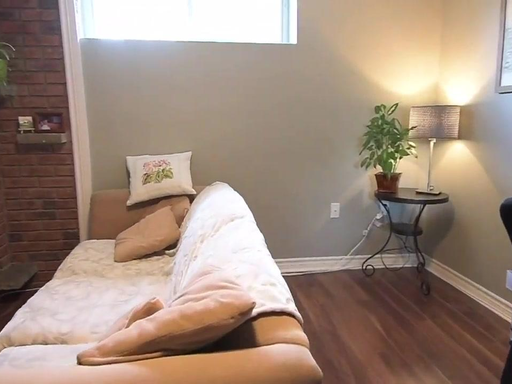} &
    \includegraphics[width=0.165\textwidth]{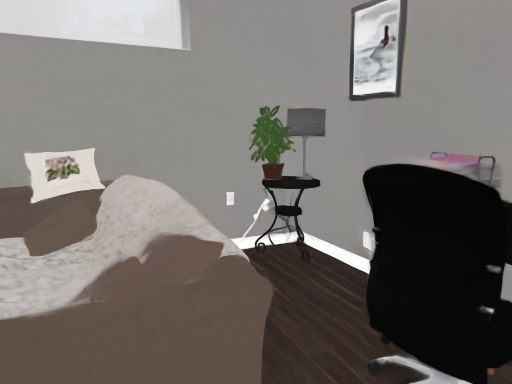} &
    \includegraphics[width=0.165\textwidth]{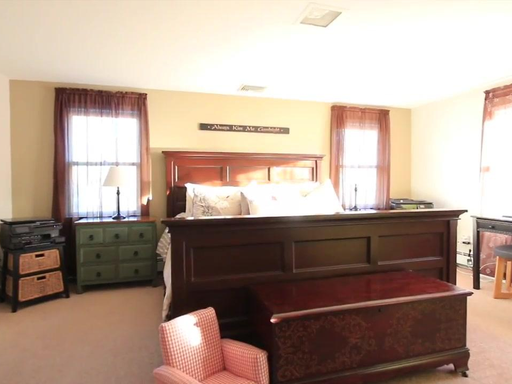} &
    \includegraphics[width=0.165\textwidth]{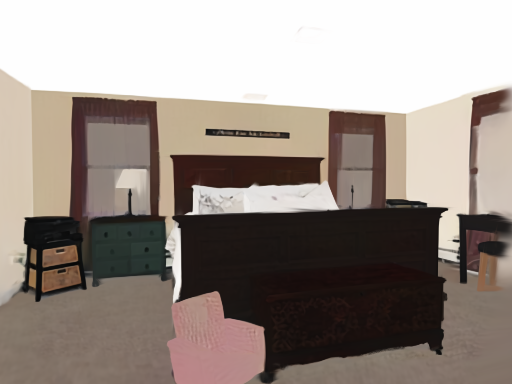} &
    \includegraphics[width=0.165\textwidth]{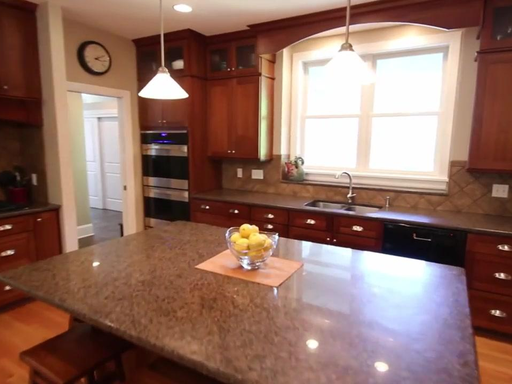} &
    \includegraphics[width=0.165\textwidth]{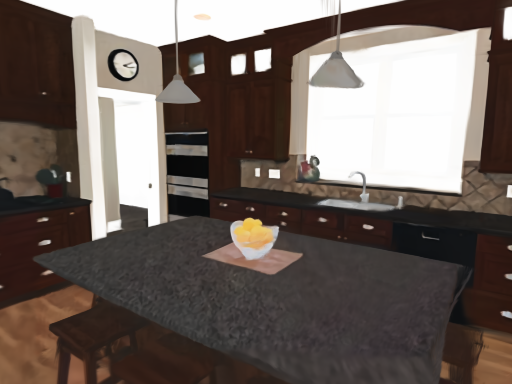} \\
  \end{tabular}%
  }
  \caption{Generalization to real-world scenes from RealEstate10K. For each of the three scenes, we show the two input views and albedo renderings at two novel views produced by our model. }\vspace{-15pt}
  \label{fig:re10k_nvs}
\end{figure*}

\begin{table}[t]
\centering
\footnotesize
\setlength{\tabcolsep}{3pt}
\caption{Multi-view consistency and albedo reconstruction on Structured3D using 2 input views per scene. * denotes fine-tuned model (on InteriorVerse). Our method achieves better multi-view consistency, especially for metallic and roughness, with the same reconstruction quality.}
\begin{tabular}{@{}lcccccccc@{}}
\toprule
\textbf{Method} & \textbf{Type} & \multicolumn{3}{c}{\textbf{Reprojection RMSE}~$\downarrow$} & \multicolumn{4}{c}{\textbf{Albedo Reconstruction}} \\
\cmidrule(lr){3-5} \cmidrule(lr){6-9}
 & & \textbf{Albedo} & \textbf{Metallic} & \textbf{Roughness} & \textbf{PSNR}$\uparrow$ & \textbf{SSIM}$\uparrow$ & \textbf{LPIPS}$\downarrow$ & \textbf{RMSE}$\downarrow$ \\
\midrule
DiffusionRenderer~\cite{diffusionrenderer} & 2D & 0.100 & 0.122 & 0.108 & 15.75 & 0.714 & 0.310 & 0.174 \\
DNF-Intrinsic~\cite{dnf_intrinsics} & 2D & 0.122 & 0.183 & 0.147 & 14.37 & 0.703 & 0.303 & 0.209 \\
MVInverse~\cite{mvinverse} & 2D & 0.044 & 0.056 & 0.038 & \second{19.83} & \second{0.771} & \second{0.268} & \second{0.108} \\
MVInverse* ~\cite{mvinverse} & 2D & \best{0.037} & \second{0.051} & \second{0.034} & \best{20.48} & \best{0.798} & \best{0.247} & \best{0.101} \\
\midrule
\textbf{Ours} & 3D & \second{0.039} & \best{0.041} & \best{0.025} & \second{19.84} & \best{0.783} & \second{0.269} & \second{0.109} \\
\bottomrule
\end{tabular}
\vspace{-16pt}
\label{tab:structured3d_consistency}
\end{table}

\begin{figure*}[!t]
  \centering
  \setlength{\tabcolsep}{0pt}
  \renewcommand{\arraystretch}{0}
  \resizebox{\textwidth}{!}{%
  \begin{tabular}{@{}>{\centering\arraybackslash}m{0.025\textwidth}@{\hspace{2pt}}*{3}{>{\centering\arraybackslash}m{0.145\textwidth}}@{\hspace{8pt}}*{3}{>{\centering\arraybackslash}m{0.145\textwidth}}@{}}
    & \multicolumn{3}{c}{\footnotesize \textbf{Ours}} & \multicolumn{3}{c}{\footnotesize \textbf{MVInverse}} \\[6pt]
    & {\footnotesize View 0} & {\footnotesize Warp 1$\to$0} & {\footnotesize Error} & {\footnotesize View 0} & {\footnotesize Warp 1$\to$0} & {\footnotesize Error} \\[6pt]
    \rotatebox{90}{\footnotesize Albedo} &
    \includegraphics[width=0.145\textwidth]{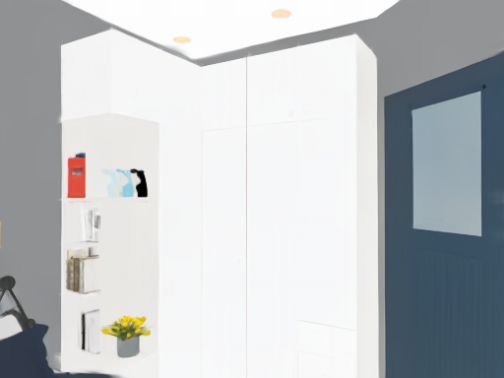} &
    \includegraphics[width=0.145\textwidth]{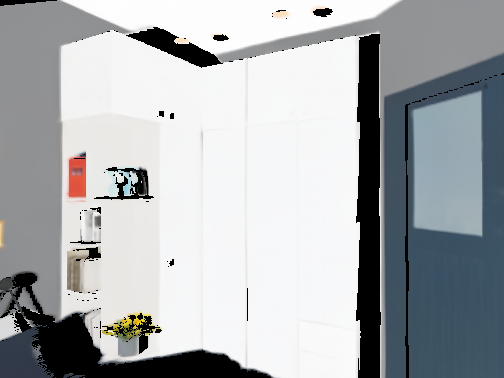} &
    \includegraphics[width=0.145\textwidth]{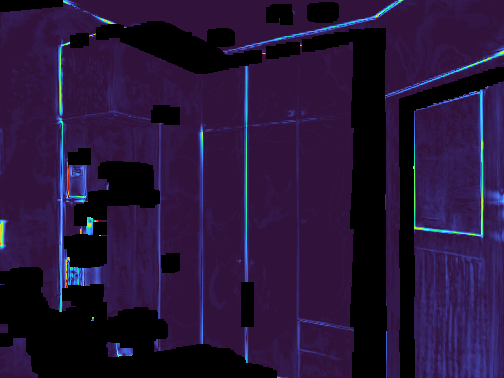} &
    \includegraphics[width=0.145\textwidth]{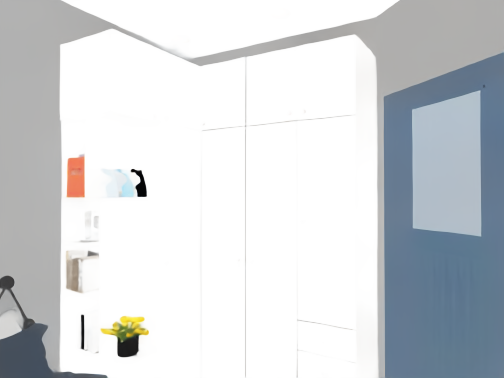} &
    \includegraphics[width=0.145\textwidth]{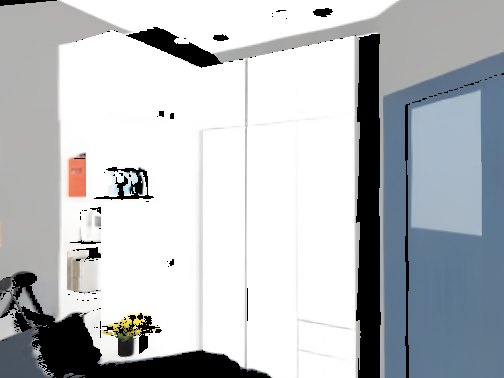} &
    \includegraphics[width=0.145\textwidth]{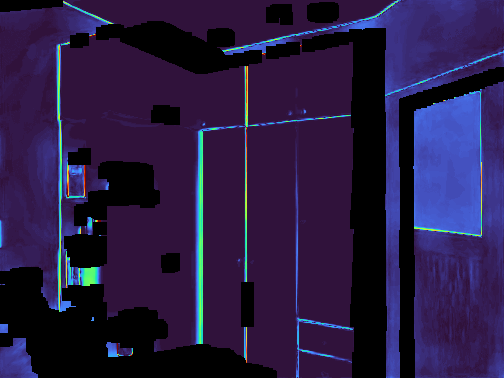} \\
    \rotatebox{90}{\footnotesize Metallic} &
    \includegraphics[width=0.145\textwidth]{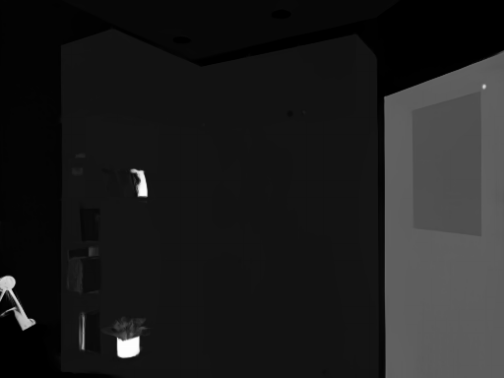} &
    \includegraphics[width=0.145\textwidth]{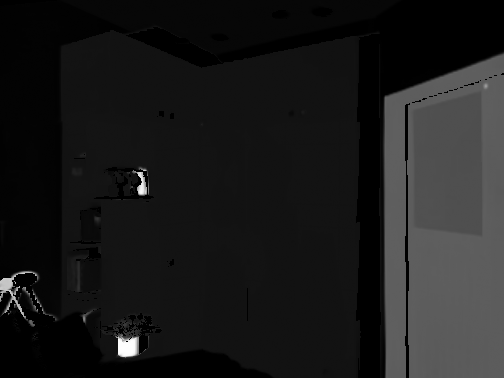} &
    \includegraphics[width=0.145\textwidth]{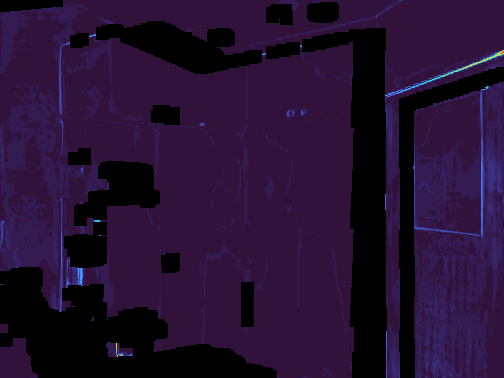} &
    \includegraphics[width=0.145\textwidth]{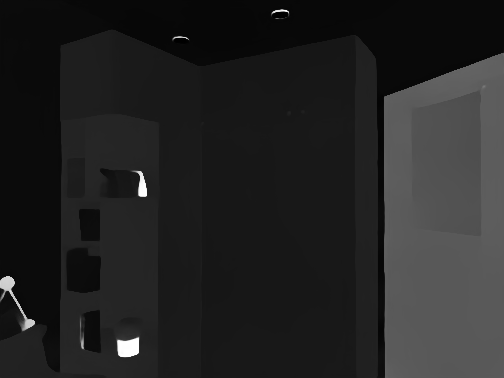} &
    \includegraphics[width=0.145\textwidth]{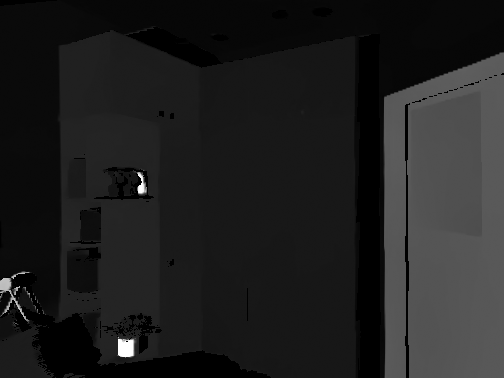} &
    \includegraphics[width=0.145\textwidth]{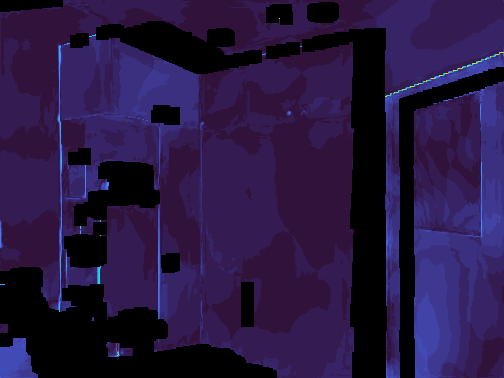} \\
    \rotatebox{90}{\footnotesize Roughness} &
    \includegraphics[width=0.145\textwidth]{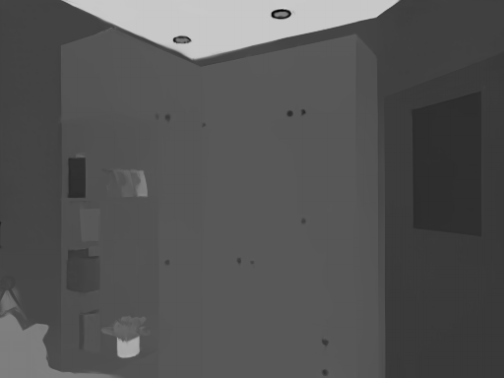} &
    \includegraphics[width=0.145\textwidth]{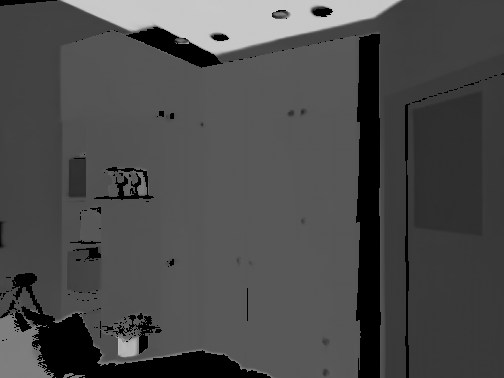} &
    \includegraphics[width=0.145\textwidth]{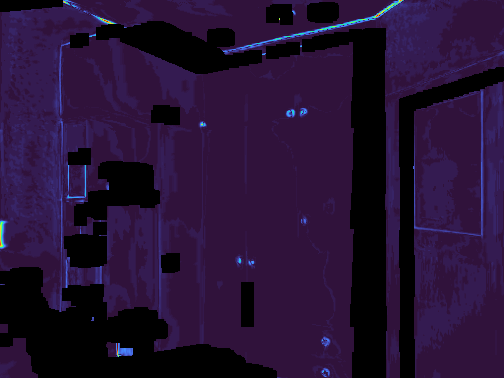} &
    \includegraphics[width=0.145\textwidth]{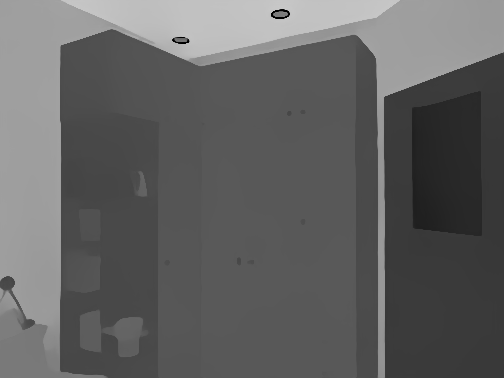} &
    \includegraphics[width=0.145\textwidth]{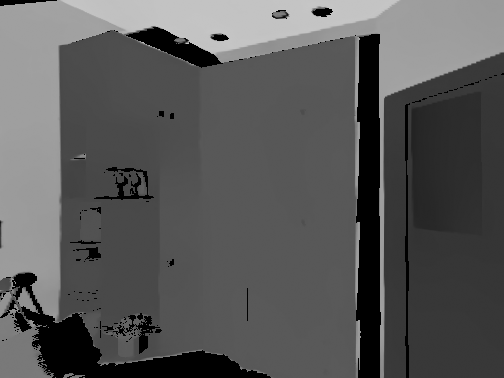} &
    \includegraphics[width=0.145\textwidth]{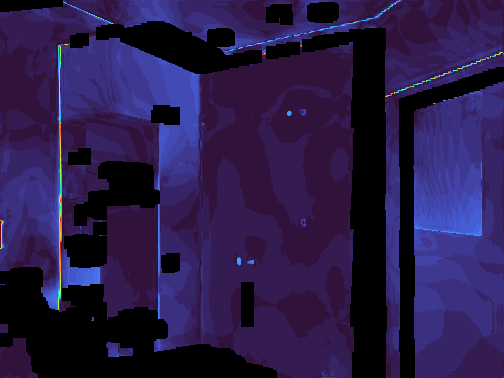} \\
    \noalign{\vskip 5pt}
    & \multicolumn{3}{c}{ \footnotesize \textbf{DiffusionRenderer}} & \multicolumn{3}{c}{\footnotesize \textbf{DNF-Intrinsic}} \\[6pt]
    & {\footnotesize View 0} & {\footnotesize Warp 1$\to$0} & {\footnotesize Error} & {\footnotesize View 0} & {\footnotesize Warp 1$\to$0} & {\footnotesize Error} \\[6pt]
    \rotatebox{90}{\footnotesize Albedo} &
    \includegraphics[width=0.145\textwidth]{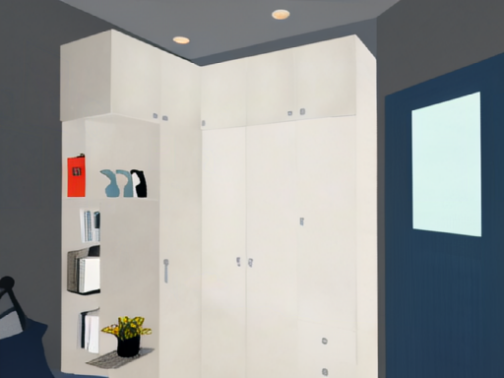} &
    \includegraphics[width=0.145\textwidth]{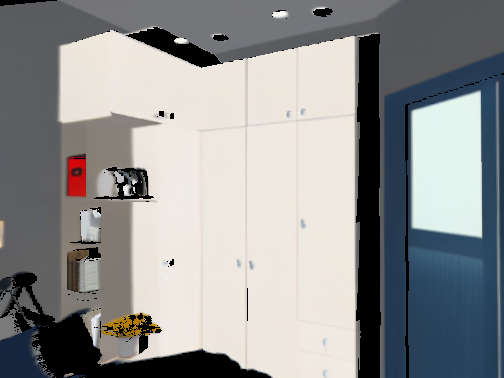} &
    \includegraphics[width=0.145\textwidth]{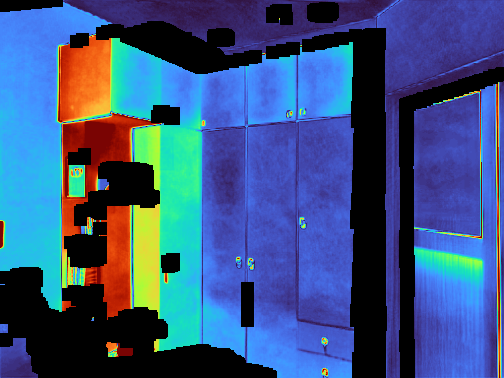} &
    \includegraphics[width=0.145\textwidth]{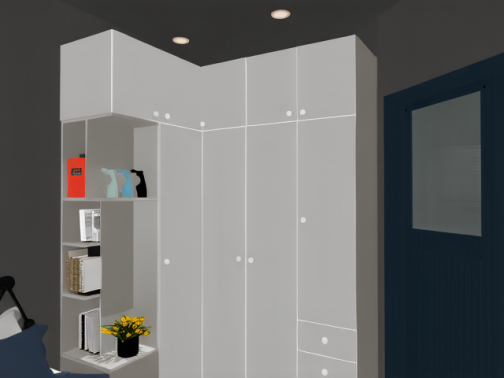} &
    \includegraphics[width=0.145\textwidth]{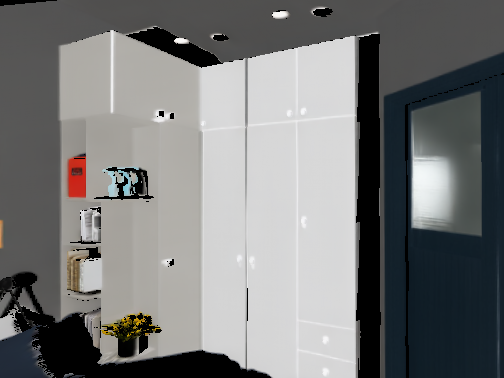} &
    \includegraphics[width=0.145\textwidth]{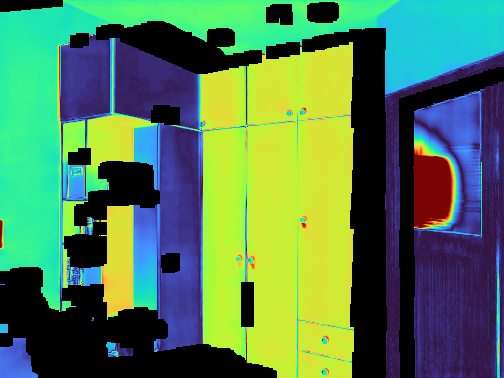} \\
    \rotatebox{90}{\footnotesize Metallic} &
    \includegraphics[width=0.145\textwidth]{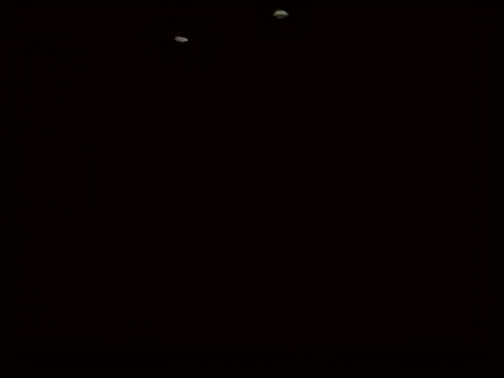} &
    \includegraphics[width=0.145\textwidth]{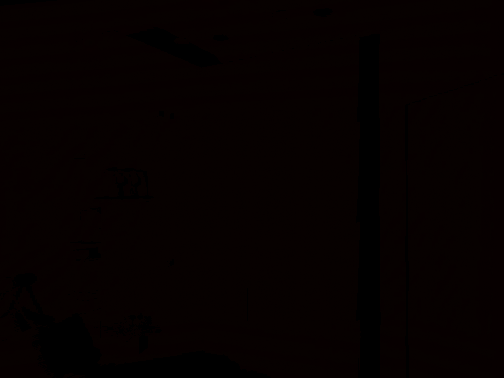} &
    \includegraphics[width=0.145\textwidth]{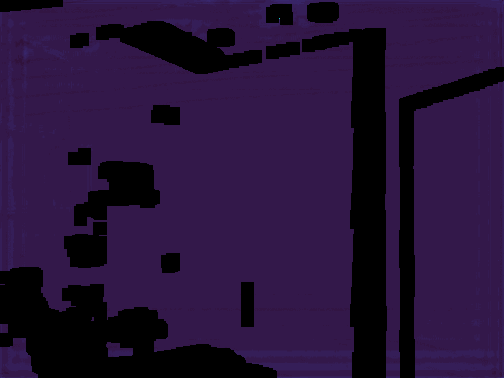} &
    \includegraphics[width=0.145\textwidth]{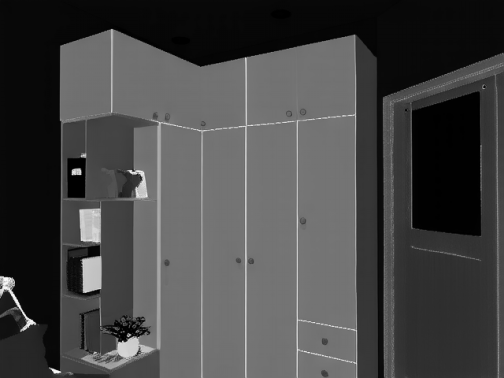} &
    \includegraphics[width=0.145\textwidth]{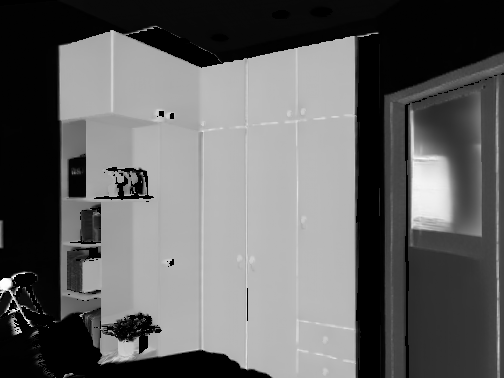} &
    \includegraphics[width=0.145\textwidth]{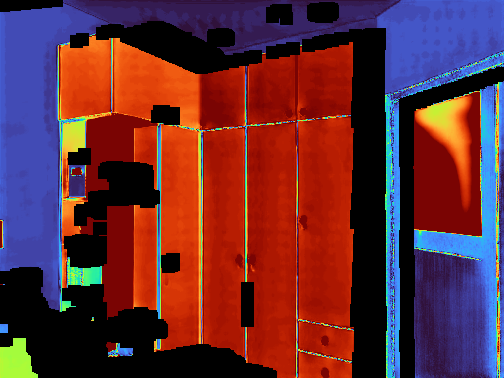} \\
    \rotatebox{90}{\footnotesize Roughness} &
    \includegraphics[width=0.145\textwidth]{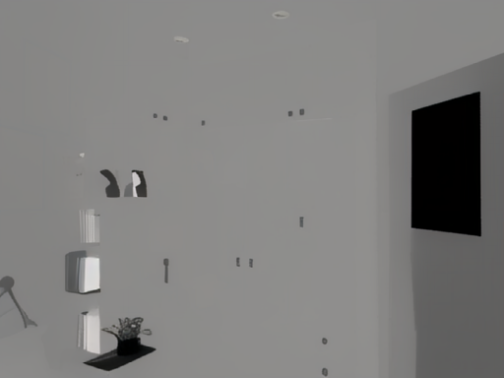} &
    \includegraphics[width=0.145\textwidth]{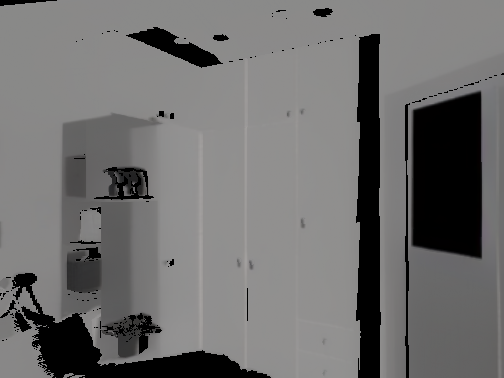} &
    \includegraphics[width=0.145\textwidth]{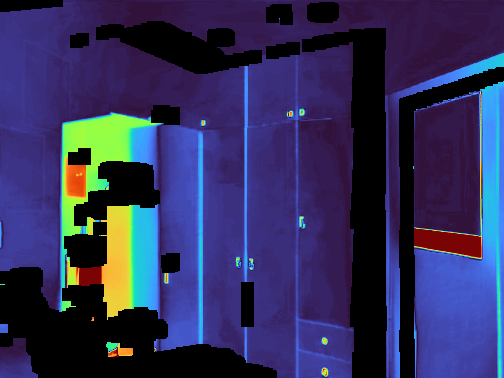} &
    \includegraphics[width=0.145\textwidth]{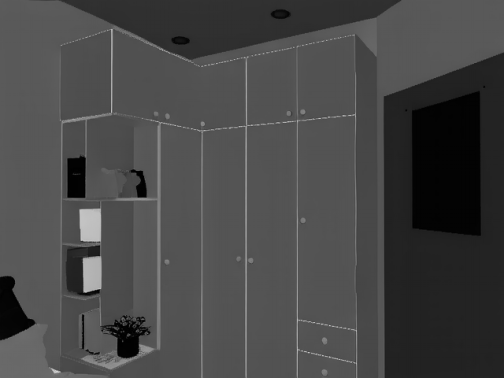} &
    \includegraphics[width=0.145\textwidth]{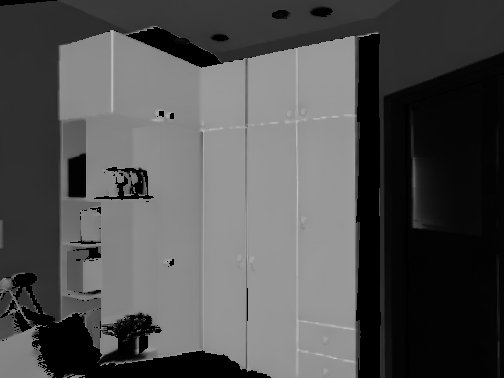} &
    \includegraphics[width=0.145\textwidth]{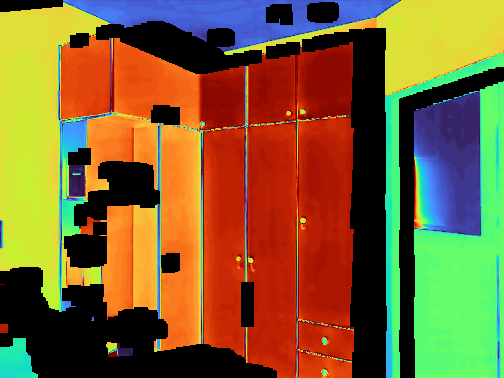} \\
  \end{tabular}%
  }
  \caption{Multi-view material consistency on a scene from Structured3D. For each method, the figure shows prediction at view 0, the prediction from view 1 warped into view 0 using ground-truth depth and pose, and the per-pixel error between them. See Supp. Figure~\ref{fig:consistency_structured3d_compressed_all} for additional examples.}\vspace{-3pt}
  \label{fig:consistency_main}
\end{figure*}

\begin{figure*}[t]
  \centering
  \setlength{\tabcolsep}{0pt}
  \renewcommand{\arraystretch}{0}
  \resizebox{\textwidth}{!}{%
  \begin{tabular}{@{} *{2}{>{\centering\arraybackslash}m{0.245\textwidth}} | *{2}{>{\centering\arraybackslash}m{0.245\textwidth}} @{}}
    \multicolumn{2}{c|}{\footnotesize Input views} & \multicolumn{2}{c}{\footnotesize Novel View Albedo} \\
    \includegraphics[width=0.245\textwidth]{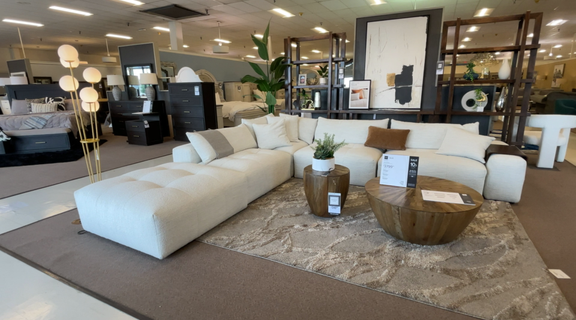} &
    \includegraphics[width=0.245\textwidth]{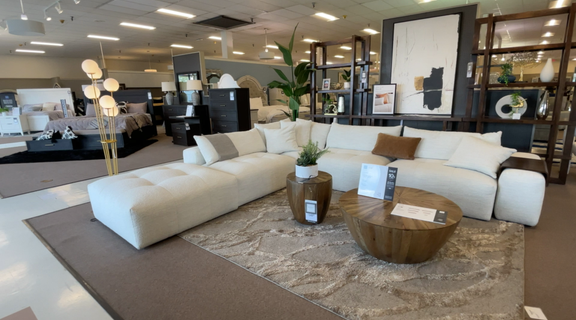} &
    \includegraphics[width=0.245\textwidth]{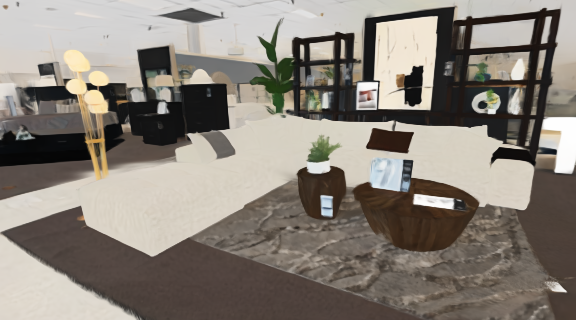} &
    \includegraphics[width=0.245\textwidth]{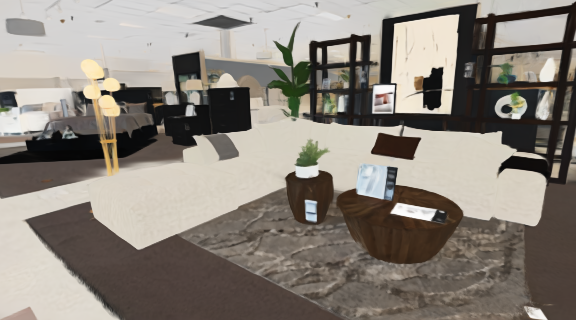} \\
    \includegraphics[width=0.245\textwidth]{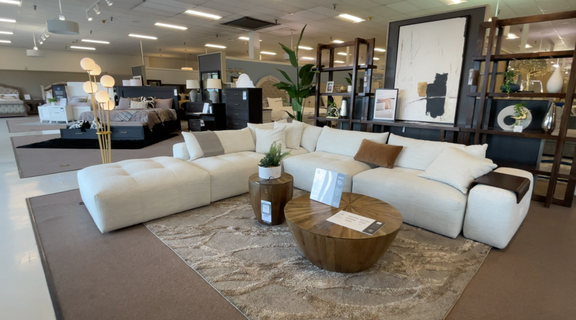} &
    \includegraphics[width=0.245\textwidth]{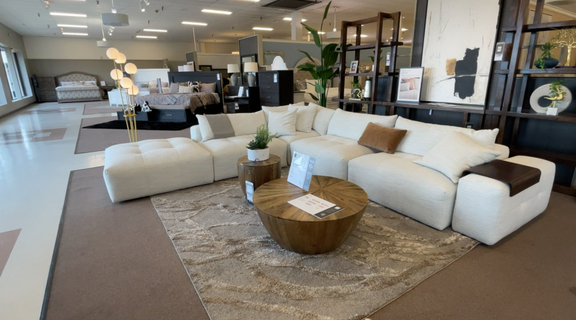} &
    \includegraphics[width=0.245\textwidth]{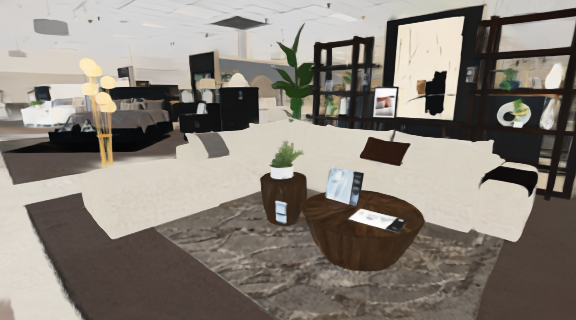} &
    \includegraphics[width=0.245\textwidth]{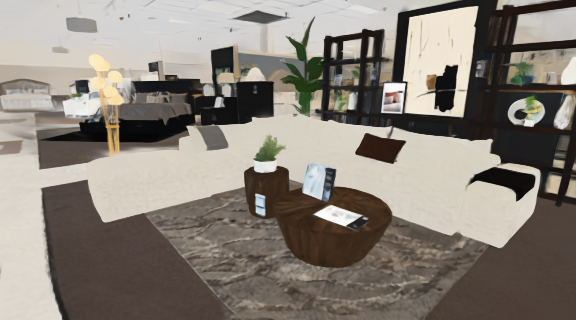} \\
  \end{tabular}%
  }
  \caption{Generalization to a real-world DL3DV scene with four input views. The first two columns show input views, and the last two columns show albedo renderings at novel viewpoints interpolated between the inputs.}
  \label{fig:dl3dv_nvs}
\end{figure*}

For \textbf{reconstruction}, we evaluate our model on the InteriorVerse dataset and report the quality of predicted materials and normals. Since albedo estimation is inherently ambiguous up to scale, we compute a per-channel scale factor that best aligns the predicted albedo with the ground-truth albedo before metric evaluation. This procedure is applied consistently across all compared methods. Quantitative results are reported in Table~\ref{tab:interiorverse_materials}. 
We also report results obtained by fine-tuning MVInverse under the same training setup as our method. 

Matching the performance of 2D pixel-aligned image networks with a unified 3D reconstruction model is inherently challenging. While our method performs slightly worse than the fine-tuned MVInverse model, it achieves performance comparable to the original MVInverse while providing an explicit 3D representation. This enables improved cross-view consistency and novel-view rendering capabilities. As shown in \Cref{fig:interiorverse_reconstruction}, our model achieves comparable reconstruction quality to 2D baselines, while avoiding baking reflections into the albedo and producing finer-grained details.

For \textbf{cross-view consistency} evaluation, we use the Structured3D dataset, which provides ground-truth albedo maps, depth maps, and camera parameters for each scene. Since the dataset does not contain smooth camera trajectories, we apply a depth-reprojection-based view selection strategy with an overlap threshold of 0.5, and evaluate on the first 251 scenes. Quantitative comparisons are reported in Table~\ref{tab:structured3d_consistency}.
We do not use InteriorVerse for consistency evaluation, as the COLMAP-estimated camera poses introduce reprojection errors that can affect the reliability of the consistency metrics.

Additionally, Figure~\ref{fig:consistency_main} visualizes the prediction from view 1 warped into view 0, together with the corresponding error maps. Existing 2D-based methods exhibit noticeable inconsistencies across views, particularly around reflective surfaces and bright highlights. In contrast, our method produces more consistent predictions due to its unified 3D Gaussian representation. The improvement is especially pronounced for metallicity and roughness, which are more challenging to infer from RGB images and benefit from stronger multi-view aggregation.

We further evaluate albedo reconstruction quality, demonstrating that our model does not simply overfit to the InteriorVerse dataset, but instead learns a more generalizable material representation.


\vspace{-3pt}
\subsection{Novel view synthesis}
We demonstrate scene reconstruction and novel-view synthesis on the RealEstate10K~\cite{zhou2018stereo} dataset in Figure~\ref{fig:re10k_nvs} and the DL3DV~\cite{ling2024dl3dv} dataset in Figure~\ref{fig:dl3dv_nvs}, with additional examples provided in the supplementary material (Figure~\ref{fig:dl3dv_nvs_example}). Our method produces plausible and view-consistent albedo decompositions on unseen real-world indoor scenes.

Although trained using only two input views, our model can generalize to a larger number of inputs at inference time, as shown in the four-view DL3DV example in Figure~\ref{fig:dl3dv_nvs} and Figure~\ref{fig:dl3dv_nvs_example_views4}. Nevertheless, artifacts may be more significant in more challenging multi-view settings with errors introduced by input poses or depth estimation.

\vspace{-3pt}
\subsection{Ablations}
\vspace{-3pt}
\begin{figure*}[t]
  \centering
  \setlength{\tabcolsep}{0pt}
  \renewcommand{\arraystretch}{0}
  \begin{tabular}{@{}ccccc@{}}
    \includegraphics[width=0.2\textwidth]{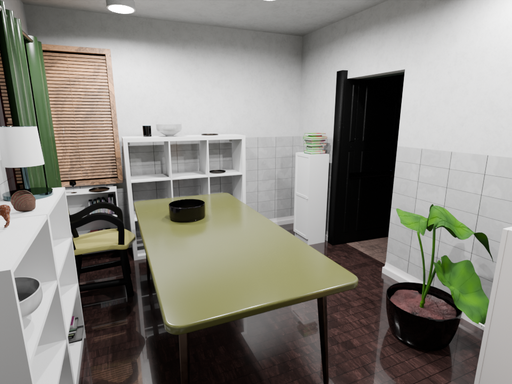} &
    \includegraphics[width=0.2\textwidth]{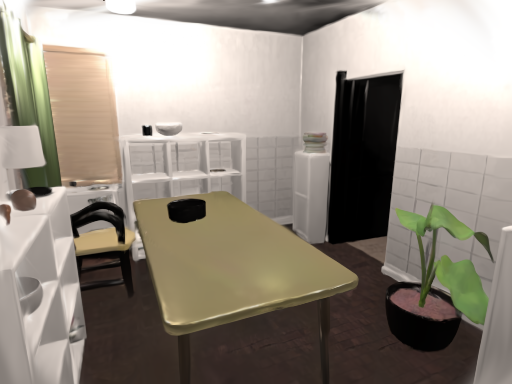} &
    \includegraphics[width=0.2\textwidth]{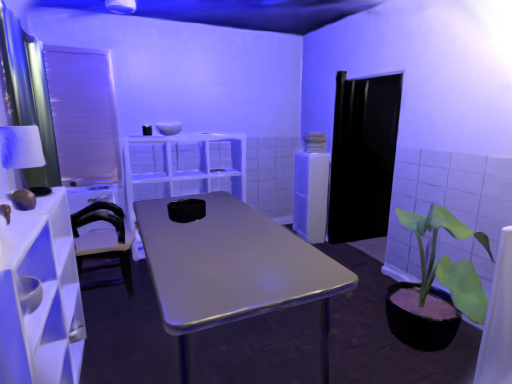} &
    \includegraphics[width=0.2\textwidth]{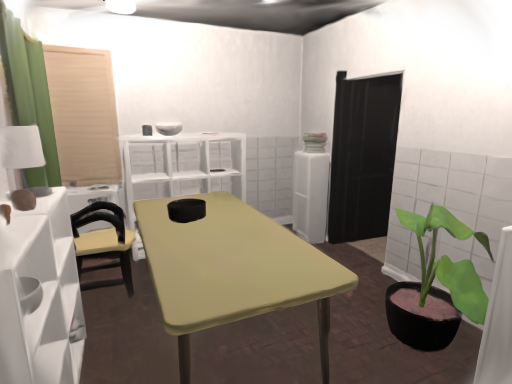} &
    \includegraphics[width=0.2\textwidth]{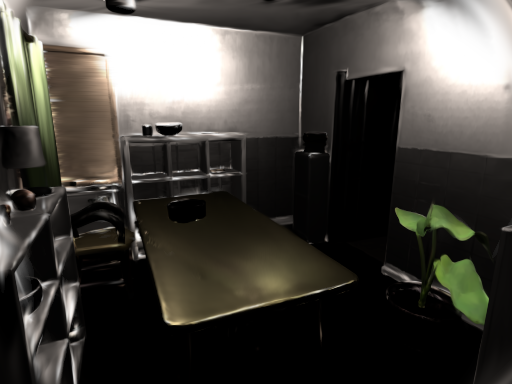} \\
    \includegraphics[width=0.2\textwidth]{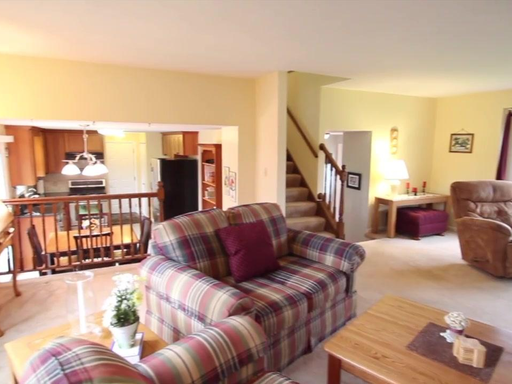} &
    \includegraphics[width=0.2\textwidth]{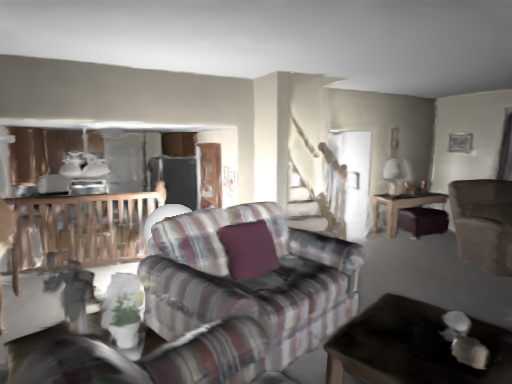} &
    \includegraphics[width=0.2\textwidth]{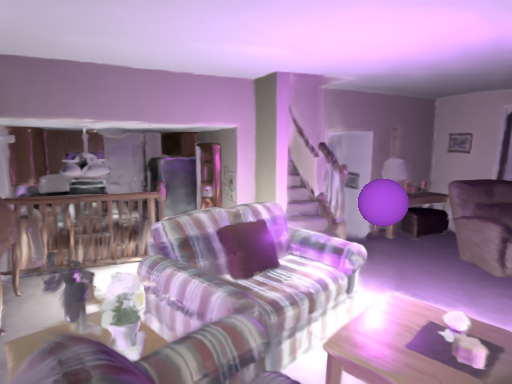} &
    \includegraphics[width=0.2\textwidth]{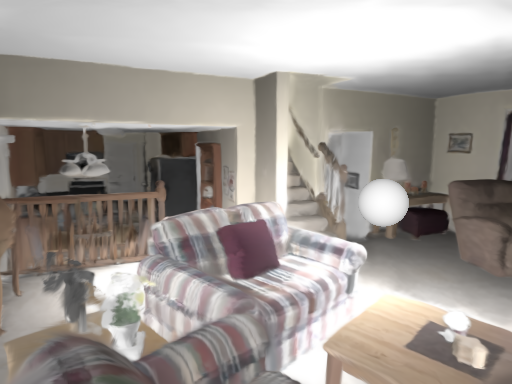} &
    \includegraphics[width=0.2\textwidth]{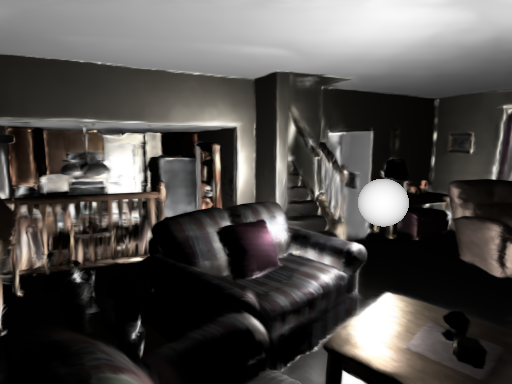}
  \end{tabular}
  \caption{ Material/lighting editing on an Infinigen scene. First row Infinigen scene, in second row RealEstate10K example. Order of images: input image (1), reconstruction under gt light for Infinigen and our selected light for RealEstate10K (2), relighting with colored lamp (3), increased roughness (4) and metallic (5) of all objects to maximum value.}
  \label{fig:infinigen_change}
  \vspace{-16pt}
\end{figure*}
\begin{table}[t]
\centering
\footnotesize
\setlength{\tabcolsep}{2.6pt}
\caption{Ablation study of 3D model design on InteriorVerse dataset using 2 input views per scene. * denotes fine-tuned network.}
\begin{tabular}{@{}lcccccccc@{}}
\toprule
\textbf{Method} & \textbf{Type} & \multicolumn{4}{c}{\textbf{Albedo}} & \textbf{Metallic} & \textbf{Roughness} & \textbf{Normal} \\
\cmidrule(lr){3-6} \cmidrule(lr){7-7} \cmidrule(lr){8-8} \cmidrule(lr){9-9}
 & & \textbf{PSNR}$\uparrow$ & \textbf{SSIM}$\uparrow$ & \textbf{LPIPS}$\downarrow$ & \textbf{RMSE}$\downarrow$ & \textbf{RMSE}$\downarrow$ & \textbf{RMSE}$\downarrow$ & \textbf{Cosine Similarity}$\uparrow$ \\
\midrule
MVInverse + Resplat* & 3D & 20.83 & 0.860 & 0.234 & 0.1011 & 0.1041 & 0.1291 & 0.9582 \\
MVInverse* + Resplat* & 3D & 21.86 & \best{0.873} & 0.212 & 0.0901 & 0.1011 & 0.1283 & 0.9607 \\
\textbf{Ours} & 3D & \best{22.18} & \best{0.873} & \best{0.203} & \best{0.0883} & \best{0.0993} & \best{0.1254} & \best{0.9609} \\
\bottomrule
\end{tabular}\vspace{-16pt}
\label{tab:naive_baselines}
\end{table}

\boldstart{Dual-branch fusion.}
As the first feed-forward framework for scene reconstruction with intrinsic materials, we additionally construct naive 3D baselines by combining existing components. Specifically, we use MVInverse~\cite{mvinverse} to predict per-view material maps and ReSplat~\cite{xu2025resplat} to predict depth and Gaussian geometry, and directly assign the predicted material maps to Gaussian primitives.
We consider two variants. In the first, MVInverse is frozen while only ReSplat is fine-tuned to predict Gaussian parameters. In the second, both MVInverse and ReSplat are jointly fine-tuned. Results are reported in Table~\ref{tab:naive_baselines}. Our unified architecture consistently outperforms both baselines across all material factors.
Furthermore, our design removes redundant backbone networks, substantially reducing model complexity while demonstrating that a single architecture can jointly solve geometry reconstruction and material estimation.

\boldstart{Normal prediction.}
We further study the design choice of explicitly predicting Gaussian normals. Among several alternatives, we find that using a dedicated normal prediction head yields the best performance. Additional details are provided in Supp.~\ref{supp:normal_ablation}.

\subsection{Applications.}
\label{sec:applications}

\boldstart{RGB reconstruction, relighting and material change.}
As an application scenario, we demonstrate that the predicted material properties can be used for relighting and material editing. To this end, we implement a simple point-light rendering simulation based on a standard BRDF shader~\cite{burley2012physically}. We evaluate this setup on synthetic scenes generated with the Infinigen~\cite{infinigen2024indoors} framework, as well as on real-world scenes from RealEstate10K. 
As shown in \Cref{fig:infinigen_change}, our method enables realistic modifications of scene appearance under novel lighting conditions. Such capabilities are valuable for downstream applications including movies, games, and interactive virtual environments, where relighting under spatially varying illumination is essential.

\boldstart{Potential for feed-forward scene reconstruction methods.}
Our representation also highlights the potential of intrinsic-material-based scene representations for future feed-forward reconstruction methods. Existing feed-forward approaches typically model appearance using spherical harmonics coefficients attached to each Gaussian primitive. In practice, these representations often smooth out or underfit high-frequency view-dependent effects such as specular highlights, resulting in appearance closer to view-independent RGB colors.
In contrast, our method explicitly predicts physically meaningful, view-independent material parameters that can be rendered under novel illumination in a physically grounded manner. This representation provides greater flexibility and stronger potential for high-quality RGB reconstruction and relighting. We support this observation with examples on both synthetic scenes (Figure~\ref{fig:relighting_infinigen}) and real-world scenes (Figure~\ref{fig:relighting_real}).


\section{Conclusion}
We present \ourmethod, a feed-forward model for scene reconstruction with material estimation from multi-view images. 
By estimating material parameters (albedo, roughness, metallicity) together with the geometry of each Gaussian primitive, our method achieves high-quality, view-consistent materials and enables novel-view synthesis of material maps. 
Experimental results on synthetic and real-world data demonstrate strong performance and support applications such as relighting and material editing.

\begin{small}
    \bibliographystyle{unsrt}
    \bibliography{refs}

@String(CVPR  = {IEEE Conf. Comput. Vis. Pattern Recog.})

@String(ICCV  = {Int. Conf. Comput. Vis.})

@String(TOG   = {ACM Trans. Graph.})

@String(CVPR  = {CVPR})

@String(ICCV  = {ICCV})

@String(TOG   = {ACM TOG})

@inproceedings{iris,
  author    = {Lin, Chih-Hao and Huang, Jia-Bin and Li, Zhengqin and Dong, Zhao and Richardt, Christian and Li, Tuotuo and Zollh{\"o}fer, Michael and Kopf, Johannes and Wang, Shenlong and Kim, Changil},
  title     = {IRIS: Inverse Rendering of Indoor Scenes from Low Dynamic Range Images},
  booktitle = {CVPR},
  year      = {2025},}

@inproceedings{diffusionrenderer,
  author        = {Liang, Ruofan and Gojcic, Zan and Ling, Huan and Munkberg, Jacob and Hasselgren, Jon and Lin, Zhi-Hao and Gao, Jun and Keller, Alexander and Vijaykumar, Nandita and Fidler, Sanja and Wang, Zian},
  title         = {DiffusionRenderer: Neural Inverse and Forward Rendering with Video Diffusion Models},
  booktitle     = {Proceedings of the IEEE/CVF Conference on Computer Vision and Pattern Recognition (CVPR)},
  year          = {2025},
  month         = jun,
  eprint        = {2501.18590},
  archivePrefix = {arXiv},
  primaryClass  = {cs.CV},
  url           = {https://research.nvidia.com/labs/toronto-ai/DiffusionRenderer/}
}

@article{dnf_intrinsics,
  author        = {Zheng, Rongjia and Zhang, Qing and Long, Chengjiang and Zheng, Wei-Shi},
  title         = {DNF-Intrinsic: Deterministic Noise-Free Diffusion for Indoor Inverse Rendering},
  journal       = {arXiv preprint arXiv:2507.03924},
  year          = {2025},
  note          = {Accepted to ICCV 2025},
  doi           = {10.48550/arXiv.2507.03924},
  url           = {https://arxiv.org/abs/2507.03924v2}
}

@article{mvinverse,
  author  = {Wu, Xiangzuo and Ren, Chengwei and Zhou, Jun and Li, Xiu and Liu, Yuan},
  title   = {MVInverse: Feed-forward Multi-view Inverse Rendering in Seconds},
  journal = {arXiv preprint arXiv:2512.21003},
  year    = {2025},
  url     = {https://maddog241.github.io/mvinverse-page/}
}

@inproceedings{intrinsicimagefusion,
  author    = {Kocsis, Peter and H{\"o}llein, Lukas and Nie{\ss}ner, Matthias},
  title     = {Intrinsic Image Fusion for Multi-View 3D Material Reconstruction},
  booktitle = {Proceedings of the IEEE/CVF Conference on Computer Vision and Pattern Recognition (CVPR)},
  year      = {2026},
  url       = {https://peter-kocsis.github.io/IntrinsicImageFusion/}
}

@article{worldmirror,
  author        = {Liu, Yifan and Min, Zhiyuan and Wang, Zhenwei and Wu, Junta and Wang, Tengfei and Yuan, Yixuan and Luo, Yawei and Guo, Chunchao},
  title         = {WorldMirror: Universal 3D World Reconstruction with Any-Prior Prompting},
  journal       = {arXiv preprint arXiv:2510.10726},
  year          = {2025},
  doi           = {10.48550/arXiv.2510.10726},
  url           = {https://arxiv.org/abs/2510.10726},
  archivePrefix = {arXiv},
  eprint        = {2510.10726},
  primaryClass  = {cs.CV}
}

@article{kerbl20233d,
  title={3d gaussian splatting for real-time radiance field rendering.},
  author={Kerbl, Bernhard and Kopanas, Georgios and Leimk{\"u}hler, Thomas and Drettakis, George and others},
  journal={ACM Trans. Graph.},
  volume={42},
  number={4},
  pages={139--1},
  year={2023}
}

@inproceedings{azinovic2019inverse,
  title={Inverse path tracing for joint material and lighting estimation},
  author={Azinovic, Dejan and Li, Tzu-Mao and Kaplanyan, Anton and Nie{\ss}ner, Matthias},
  booktitle={Proceedings of the IEEE/CVF conference on computer vision and pattern recognition},
  pages={2447--2456},
  year={2019}
}

@article{zhang2021nerfactor,
  title={Nerfactor: Neural factorization of shape and reflectance under an unknown illumination},
  author={Zhang, Xiuming and Srinivasan, Pratul P and Deng, Boyang and Debevec, Paul and Freeman, William T and Barron, Jonathan T},
  journal={ACM Transactions on Graphics (ToG)},
  volume={40},
  number={6},
  pages={1--18},
  year={2021},
  publisher={ACM New York, NY, USA}
}

@inproceedings{boss2021nerd,
  title={Nerd: Neural reflectance decomposition from image collections},
  author={Boss, Mark and Braun, Raphael and Jampani, Varun and Barron, Jonathan T and Liu, Ce and Lensch, Hendrik},
  booktitle={Proceedings of the IEEE/CVF International Conference on Computer Vision},
  pages={12684--12694},
  year={2021}
}

@inproceedings{liang2024gs,
  title={Gs-ir: 3d gaussian splatting for inverse rendering},
  author={Liang, Zhihao and Zhang, Qi and Feng, Ying and Shan, Ying and Jia, Kui},
  booktitle={Proceedings of the IEEE/CVF Conference on Computer Vision and Pattern Recognition},
  pages={21644--21653},
  year={2024}
}

@inproceedings{li2018learning,
  title={Learning intrinsic image decomposition from watching the world},
  author={Li, Zhengqi and Snavely, Noah},
  booktitle={Proceedings of the IEEE conference on computer vision and pattern recognition},
  pages={9039--9048},
  year={2018}
}

@inproceedings{li2020inverse,
  title={Inverse rendering for complex indoor scenes: Shape, spatially-varying lighting and svbrdf from a single image},
  author={Li, Zhengqin and Shafiei, Mohammad and Ramamoorthi, Ravi and Sunkavalli, Kalyan and Chandraker, Manmohan},
  booktitle={Proceedings of the IEEE/CVF conference on computer vision and pattern recognition},
  pages={2475--2484},
  year={2020}
}

@inproceedings{zhu2022learning,
  title={Learning-based inverse rendering of complex indoor scenes with differentiable monte carlo raytracing},
  author={Zhu, Jingsen and Luan, Fujun and Huo, Yuchi and Lin, Zihao and Zhong, Zhihua and Xi, Dianbing and Wang, Rui and Bao, Hujun and Zheng, Jiaxiang and Tang, Rui},
  booktitle={Siggraph asia 2022 conference papers},
  pages={1--8},
  year={2022}
}

@article{depthanything3,
  title={Depth Anything 3: recovering the visual space from any views},
  author={Haotong Lin and Sili Chen and Jun Hao Liew and Donny Y. Chen and Zhenyu Li and Guang Shi and Jiashi Feng and Bingyi Kang},
  journal={arXiv preprint arXiv:2511.10647},
  year={2025}
}

@inproceedings{wang2025vggt,
  title={VGGT: Visual Geometry Grounded Transformer},
  author={Wang, Jianyuan and Chen, Minghao and Karaev, Nikita and Vedaldi, Andrea and Rupprecht, Christian and Novotny, David},
  booktitle={Proceedings of the IEEE/CVF Conference on Computer Vision and Pattern Recognition},
  year={2025}
}

@InProceedings{Wang_2024_CVPR,
    author    = {Wang, Shuzhe and Leroy, Vincent and Cabon, Yohann and Chidlovskii, Boris and Revaud, Jerome},
    title     = {DUSt3R: Geometric 3D Vision Made Easy},
    booktitle = {Proceedings of the IEEE/CVF Conference on Computer Vision and Pattern Recognition (CVPR)},
    month     = {June},
    year      = {2024},
    pages     = {20697-20709}
}

@article{xu2025resplat,
  title={Resplat: Learning recurrent gaussian splats},
  author={Xu, Haofei and Barath, Daniel and Geiger, Andreas and Pollefeys, Marc},
  journal={arXiv preprint arXiv:2510.08575},
  year={2025}
}

@inproceedings{xu2025depthsplat,
  title={Depthsplat: Connecting gaussian splatting and depth},
  author={Xu, Haofei and Peng, Songyou and Wang, Fangjinhua and Blum, Hermann and Barath, Daniel and Geiger, Andreas and Pollefeys, Marc},
  booktitle={Proceedings of the Computer Vision and Pattern Recognition Conference},
  pages={16453--16463},
  year={2025}
}

@inproceedings{charatan2024pixelsplat,
  title={pixelsplat: 3d gaussian splats from image pairs for scalable generalizable 3d reconstruction},
  author={Charatan, David and Li, Sizhe Lester and Tagliasacchi, Andrea and Sitzmann, Vincent},
  booktitle={Proceedings of the IEEE/CVF conference on computer vision and pattern recognition},
  pages={19457--19467},
  year={2024}
}

@inproceedings{chen2024mvsplat,
  title={Mvsplat: Efficient 3d gaussian splatting from sparse multi-view images},
  author={Chen, Yuedong and Xu, Haofei and Zheng, Chuanxia and Zhuang, Bohan and Pollefeys, Marc and Geiger, Andreas and Cham, Tat-Jen and Cai, Jianfei},
  booktitle={European conference on computer vision},
  pages={370--386},
  year={2024},
  organization={Springer}
}

@article{ye2024no,
  title={No pose, no problem: Surprisingly simple 3d gaussian splats from sparse unposed images},
  author={Ye, Botao and Liu, Sifei and Xu, Haofei and Li, Xueting and Pollefeys, Marc and Yang, Ming-Hsuan and Peng, Songyou},
  journal={arXiv preprint arXiv:2410.24207},
  year={2024}
}

@article{oquab2023dinov2,
  title={Dinov2: Learning robust visual features without supervision},
  author={Oquab, Maxime and Darcet, Timoth{\'e}e and Moutakanni, Th{\'e}o and Vo, Huy and Szafraniec, Marc and Khalidov, Vasil and Fernandez, Pierre and Haziza, Daniel and Massa, Francisco and El-Nouby, Alaaeldin and others},
  journal={arXiv preprint arXiv:2304.07193},
  year={2023}
}

@article{Ranftl2020,
	author    = {Ren\'{e} Ranftl and Katrin Lasinger and David Hafner and Konrad Schindler and Vladlen Koltun},
	title     = {Towards Robust Monocular Depth Estimation: Mixing Datasets for Zero-shot Cross-dataset Transfer},
	journal   = {IEEE Transactions on Pattern Analysis and Machine Intelligence (TPAMI)},
	year      = {2020},
}

@inproceedings{zhao2021point,
  title={Point transformer},
  author={Zhao, Hengshuang and Jiang, Li and Jia, Jiaya and Torr, Philip HS and Koltun, Vladlen},
  booktitle={Proceedings of the IEEE/CVF international conference on computer vision},
  pages={16259--16268},
  year={2021}
}

@inproceedings{schonberger2016structure,
  title={Structure-from-motion revisited},
  author={Schonberger, Johannes L and Frahm, Jan-Michael},
  booktitle={Proceedings of the IEEE conference on computer vision and pattern recognition},
  pages={4104--4113},
  year={2016}
}

@inproceedings{zheng2020structured3d,
  title={Structured3d: A large photo-realistic dataset for structured 3d modeling},
  author={Zheng, Jia and Zhang, Junfei and Li, Jing and Tang, Rui and Gao, Shenghua and Zhou, Zihan},
  booktitle={European Conference on Computer Vision},
  pages={519--535},
  year={2020},
  organization={Springer}
}

@inproceedings{infinigen2024indoors,
    author    = {Raistrick, Alexander and Mei, Lingjie and Kayan, Karhan and Yan, David and Zuo, Yiming and Han, Beining and Wen, Hongyu and Parakh, Meenal and Alexandropoulos, Stamatis and Lipson, Lahav and Ma, Zeyu and Deng, Jia},
    title     = {Infinigen Indoors: Photorealistic Indoor Scenes using Procedural Generation},
    booktitle = {Proceedings of the IEEE/CVF Conference on Computer Vision and Pattern Recognition (CVPR)},
    month     = {June},
    year      = {2024},
    pages     = {21783-21794}
}

@article{zhou2018stereo,
  title={Stereo magnification: Learning view synthesis using multiplane images},
  author={Zhou, Tinghui and Tucker, Richard and Flynn, John and Fyffe, Graham and Snavely, Noah},
  journal={arXiv preprint arXiv:1805.09817},
  year={2018}
}

@inproceedings{he2015deep,
  title={Deep residual learning for image recognition. 2016 IEEE Conf},
  author={He, Kaiming and Zhang, Xiangyu and Ren, Shaoquing and Sun, Jian},
  booktitle={Comput. Vis. Pattern Recognit},
  pages={770--778},
  year={2015}
}

@inproceedings{wang2025moge,
  title={Moge: Unlocking accurate monocular geometry estimation for open-domain images with optimal training supervision},
  author={Wang, Ruicheng and Xu, Sicheng and Dai, Cassie and Xiang, Jianfeng and Deng, Yu and Tong, Xin and Yang, Jiaolong},
  booktitle={Proceedings of the Computer Vision and Pattern Recognition Conference},
  pages={5261--5271},
  year={2025}
}

@inproceedings{ling2024dl3dv,
  title={Dl3dv-10k: A large-scale scene dataset for deep learning-based 3d vision},
  author={Ling, Lu and Sheng, Yichen and Tu, Zhi and Zhao, Wentian and Xin, Cheng and Wan, Kun and Yu, Lantao and Guo, Qianyu and Yu, Zixun and Lu, Yawen and others},
  booktitle={Proceedings of the IEEE/CVF Conference on Computer Vision and Pattern Recognition},
  pages={22160--22169},
  year={2024}
}

@article{parker2010optix,
  title={Optix: a general purpose ray tracing engine},
  author={Parker, Steven G and Bigler, James and Dietrich, Andreas and Friedrich, Heiko and Hoberock, Jared and Luebke, David and McAllister, David and McGuire, Morgan and Morley, Keith and Robison, Austin and others},
  journal={Acm transactions on graphics (tog)},
  volume={29},
  number={4},
  pages={1--13},
  year={2010},
  publisher={ACM New York, NY, USA}
}

@inproceedings{burley2012physically,
  title={Physically-based shading at disney},
  author={Burley, Brent and Studios, Walt Disney Animation},
  booktitle={Acm siggraph},
  volume={2012},
  number={2012},
  pages={1--7},
  year={2012},
  organization={vol. 2012}
}
\end{small}

\appendix
\newpage
\section{Supplementary}

\subsection{Architecture details}
\label{supp:architecture}
Although the geometry and intrinsic branches of \ourmethod\ each originate from a different pretrained network --- ReSplat~\cite{xu2025resplat} and MVInverse~\cite{mvinverse}, respectively --- the two branches are not run independently. Because the decoder heads see features different from those its pretrained version was trained on, the performance of each of them is initially degraded; joint end-to-end fine-tuning then recovers it.

\paragraph{Translator features feed the depth pipeline.}
The four features from different layers $\{\mathbf{F}^m_\ell\}_{\ell\in\{3,8,13,17\}}$ produced by the Multi-view Intrinsic Translator are projected by linear layers and concatenated with the geometry-branch features as the ``mono'' input of the depth pipeline, replacing the multi-scale features that ReSplat's original depth pipeline takes from intermediate layers of its own DINOv2 backbone.

\paragraph{Geometry ResNet feeds the Albedo head.}
MVInverse originally fuses a separate frozen ResNeXt-101 image pyramid as multi-scale residual features into its albedo DPT head. We remove this ResNeXt entirely and feed the geometry branch's ResNet pyramid into the albedo head instead, through a learned $1\!\times\!1$ convolutional channel adaptor that matches the ResNet channel widths to those expected by the DPT. The metallicity, roughness, and normal heads operate on translator features only and add no further coupling.

\paragraph{Depth head.}
Following ReSplat, the depth head is a UNet whose input concatenates the cost volume $\mathbf{C}_i$, the multi-scale ResNet pyramid, the Multi-view Geometry Encoder features, and the translator features. The UNet regresses a softmax distribution over a fixed set of inverse-depth candidates, from which we read out a per-pixel depth $d_i$; a DPT residual module then upsamples this depth map to input resolution.

\paragraph{Material and Normal heads.}
Following MVInverse, the three material heads are DPT decoders that consume the four translator features $\{\mathbf{F}^m_\ell\}$. The albedo head additionally fuses the multi-scale ResNet pyramid as residual evidence via the channel adaptor described above, while the metallicity and roughness heads use translator features alone.

\paragraph{Gaussian shape head.}
The Gaussian shape head, following ReSplat, predicts the per-primitive rotation $\mathbf{q}_j$, scale $\mathbf{s}_j$, and opacity $\sigma_j$. Its input is a per-pixel feature concatenation comprising the unshuffled image patches, the predicted depth, the cost-volume match probability, and the fused multi-scale features; this representation is lifted to 3D using the predicted depth and refined with a Point Transformer~\cite{zhao2021point} that aggregates information from $k$-nearest spatial neighbours. A small MLP then maps the resulting 3D-aware features to the per-primitive parameters.

\subsection{Implementation Details}
\paragraph{Dataset.} InteriorVerse provides ground-truth material maps (albedo/metallic/roughness) and geometry maps (depth/normal). Since it does not provide camera poses and views are sampled randomly within scenes (rather than along a smooth video trajectory), we first estimated camera poses using COLMAP~\cite{schonberger2016structure}, and then selected views with large overlap (>0.4) via depth reprojection.
As the dataset is very sparse (between 2 and 80 views per scene) and indoor scenes also contain large textureless regions, we found COLMAP estimation can be noisy and incomplete. To mitigate this, we ran ReSplat to select triplets with good RGB reconstruction quality (PSNR > 23) for training. For testing, we used one pair per scene with the largest overlap, without any ReSplat filtering, to avoid introducing selection bias when comparing to 2D methods. We followed the InteriorVerse train/test split. After all filtering, we obtained 13k triplets for training and 137 pairs for testing.

Following several of our baselines, as InteriorVerse images contain noise from dataset rendering, we run the OptiX~\cite{parker2010optix} denoiser on the input RGB images and use tone-mapping. We found that each baseline performs better with either the original input or the denoised input, and we report the best results for each method.

\paragraph{Training details.}
All models are trained with AdamW (weight decay 0.01) using linear warmup followed by cosine annealing. For the MVInverse fine-tuning experiment, we fine-tune the DINOv2 backbone, decoder, and the normal and material heads with $\text{LR}=1\mathrm{e}{-5}$ and 500 warmup steps. For our 3D experiments, we train the DINOv2 backbone and the Multi-view Intrinsic Translator with $\text{LR}=1\mathrm{e}{-5}$ and 1000 warmup steps, and the remaining components with $\text{LR}=1\mathrm{e}{-6}$ and 1000 warmup steps.
For the loss weights, we use $w=1.0$ for the $L_1$ albedo loss, $w=0.5$ for the $L_1$ metallic and $L_1$ roughness losses, $w=1.0$ for the $L_1$ MoGe-style~\cite{wang2025moge} depth loss, $w=1.0$ for the $L_1$ normal consistency loss, and $w=1.0$ for the per-material LPIPS losses on albedo, metallic, and roughness. As metallic and roughness maps are one-channel, we replicate them three times to compute LPIPS. 

\subsection{Normal ablation.}
\label{supp:normal_ablation}
\begin{wrapfigure}[20]{r}{0.4\linewidth}
\centering
\setlength{\tabcolsep}{0pt}
\renewcommand{\arraystretch}{0}
\resizebox{\linewidth}{!}{%
\begin{tabular}{@{}cc@{}}
  \includegraphics[width=0.48\linewidth]{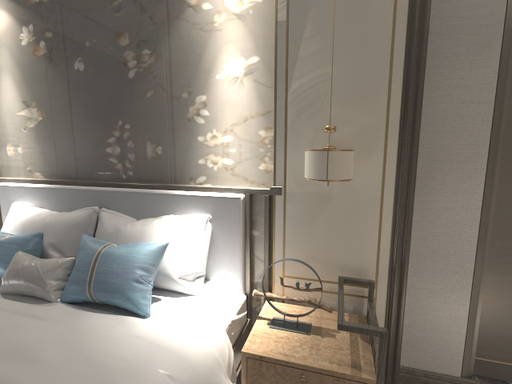} &
  \includegraphics[width=0.48\linewidth]{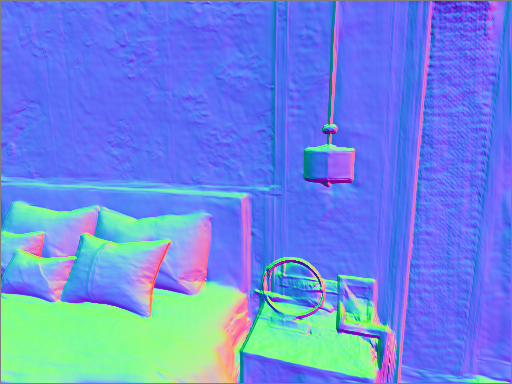} \\[4pt]
  \small Input & \small Normals from depth \\[8pt]
  \includegraphics[width=0.48\linewidth]{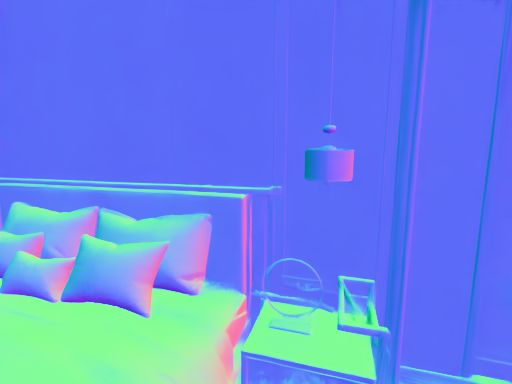} &
  \includegraphics[width=0.48\linewidth]{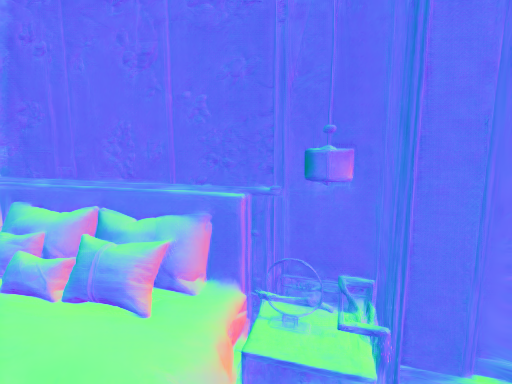} \\[4pt]
  \small Normal Head (Ours) & \small Gaussian Head \\[8pt]
\end{tabular}
}
\caption{Alternative derivation for gaussian normals. First row: input image, normals derived from depth. Second row: rendered normals, left is separate head for prediction, right is prediction in gaussian head.}
\label{fig:normal_ablation}
\end{wrapfigure}
We also ablate the normal prediction branch and test two other variants: one in which we directly compute normals from depth using finite differences, and another in which we predict normals from the Gaussian head together with scales and opacities and then render them. In all experiments, we used a normal consistency loss with the ground truth. In both variants, the model struggles to separate geometry from appearance and leaks texture details into the normals, which harms relighting quality. Results are shown in Figure~\ref{fig:normal_ablation}. We additionally tried adding a normal smoothness loss, but this did not help much. This shows that pre-training also helps when learning Gaussian normals jointly with material factors.

\subsection{View-dependent effects modelling of feed-forward scene reconstruction methods}

Given sparse multi-view RGB images, our method reconstructs geometry together with intrinsic material factors (albedo, metallic, roughness). We then render the reconstructed Gaussians under the \emph{ground-truth} light configuration from Infinigen to produce relit RGB images, without per-scene optimization. In addition, we show that our method allows us to change the light and materials of objects.

\begin{figure*}[t]
  \centering
  \setlength{\tabcolsep}{0pt}
  \renewcommand{\arraystretch}{0}
\begin{tabular}{@{}>{\centering\arraybackslash}m{0.025\textwidth} *{5}{>{\centering\arraybackslash}m{0.19\textwidth}}@{}}
    & \textbf{GT} & \textbf{Ours} & \textbf{ReSplat} & \textbf{WorldMirror} & \textbf{AnySplat} \\[2pt]
    \rotatebox{90}{View 1} &
    \includegraphics[width=0.19\textwidth]{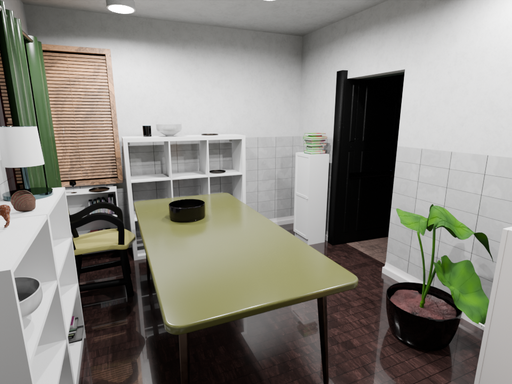} &
    \includegraphics[width=0.19\textwidth]{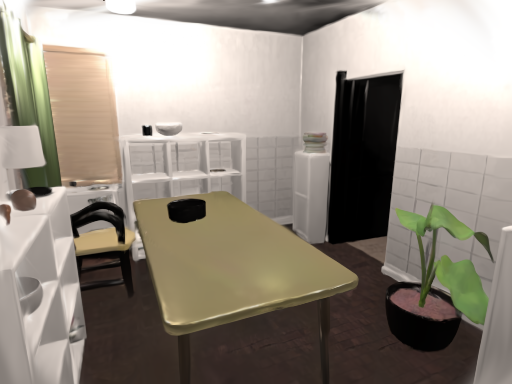} &
    \includegraphics[width=0.19\textwidth]{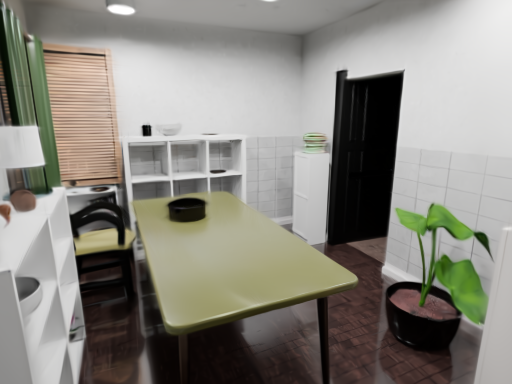} &
    \includegraphics[width=0.19\textwidth]{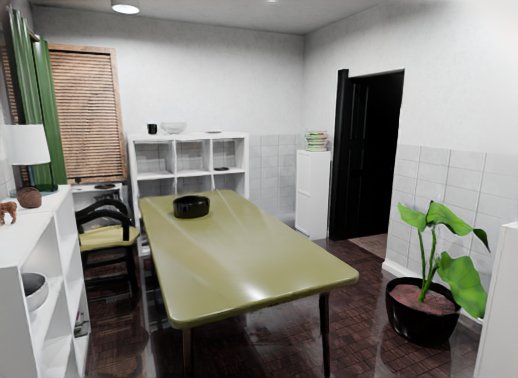} &
    \includegraphics[width=0.19\textwidth]{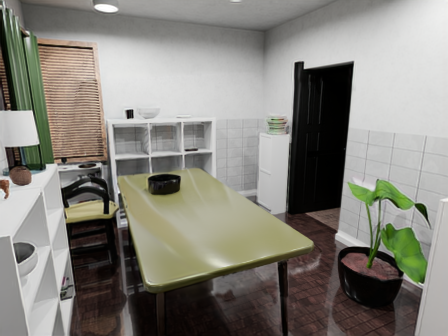} \\
    \rotatebox{90}{View 2} &
    \includegraphics[width=0.19\textwidth]{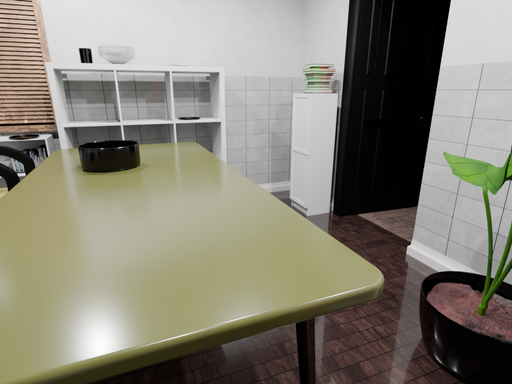} &
    \includegraphics[width=0.19\textwidth]{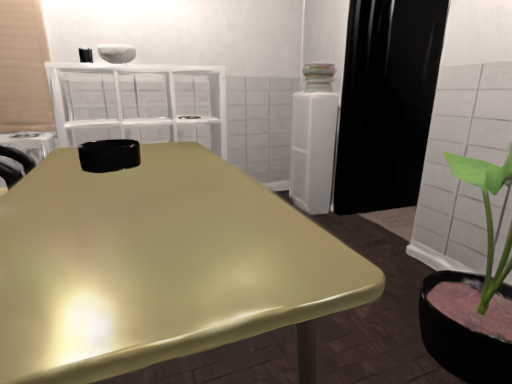} &
    \includegraphics[width=0.19\textwidth]{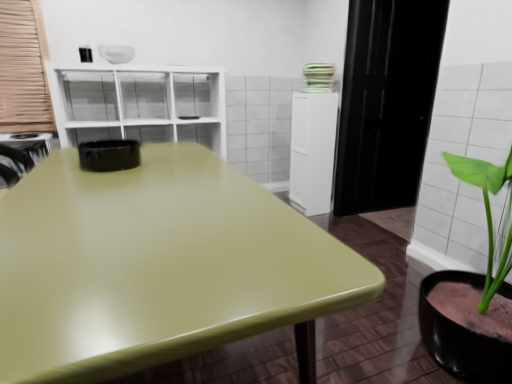} &
    \includegraphics[width=0.19\textwidth]{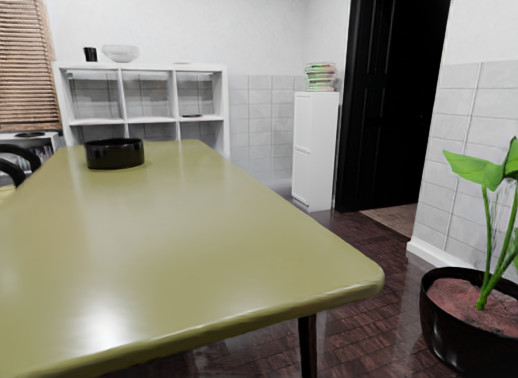} &
    \includegraphics[width=0.19\textwidth]{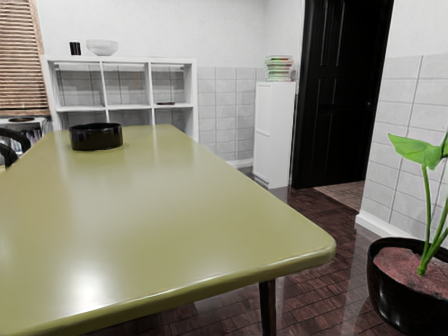} \\
  \end{tabular}
  \caption{RGB reconstruction comparison on a synthetic Infinigen scene across two views (rows) and methods (columns). Our method renders relit RGB by composing predicted material factors with the ground-truth Infinigen light configuration. For Gaussian without explicitly modeling material, the view-dependent effect is limited.}
  \label{fig:relighting_infinigen}
\end{figure*}

\begin{figure*}[t]
  \centering
  \setlength{\tabcolsep}{0pt}
  \renewcommand{\arraystretch}{0}
  \begin{tabular}{@{}>{\centering\arraybackslash}m{0.025\textwidth} *{5}{>{\centering\arraybackslash}m{0.19\textwidth}}@{}}
    & \textbf{GT} & \textbf{Ours (relight)} & \textbf{ReSplat} & \textbf{WorldMirror} & \textbf{AnySplat} \\[2pt]
    \rotatebox{90}{View 1} &
    \includegraphics[width=0.19\textwidth]{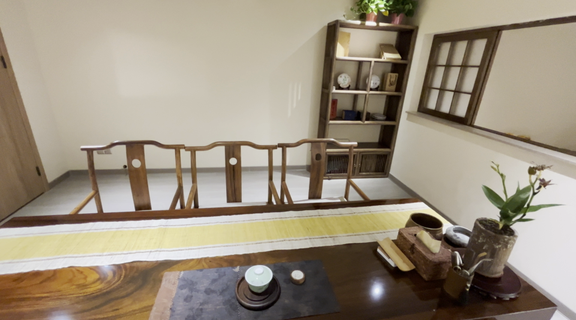} &
    \includegraphics[width=0.19\textwidth]{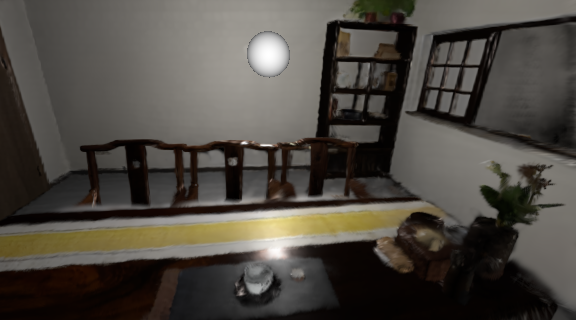} &
    \includegraphics[width=0.19\textwidth]{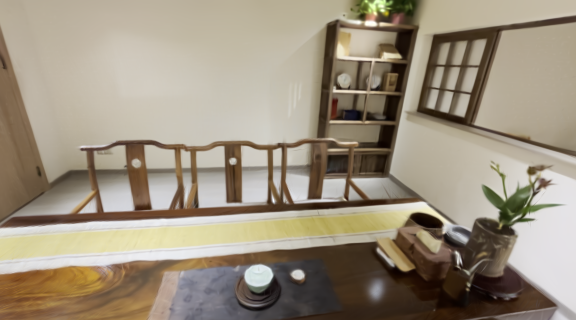} &
    \includegraphics[width=0.19\textwidth]{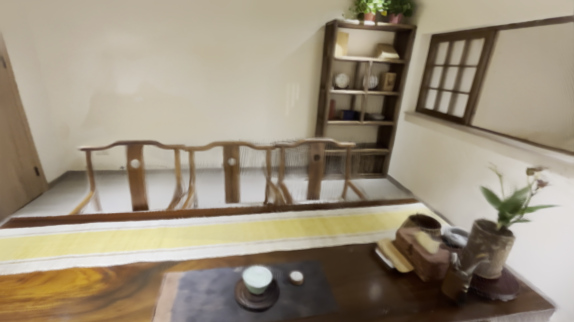} &
    \includegraphics[width=0.19\textwidth]{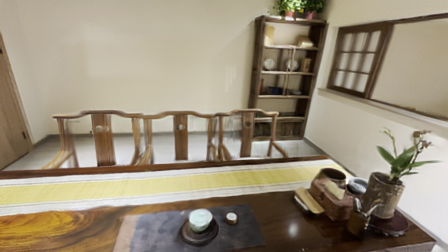} \\
    \rotatebox{90}{View 2} &
    \includegraphics[width=0.19\textwidth]{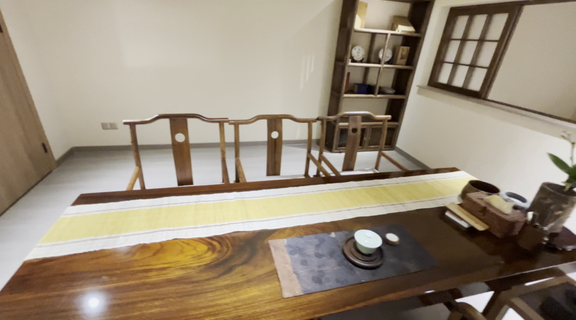} &
    \includegraphics[width=0.19\textwidth]{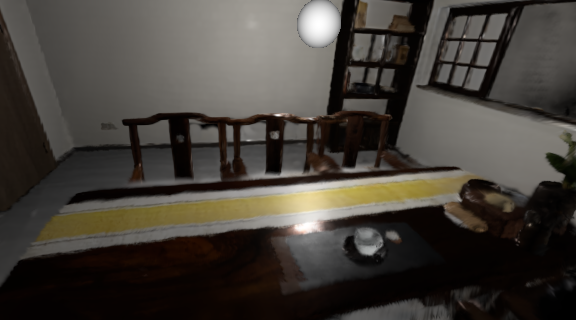} &
    \includegraphics[width=0.19\textwidth]{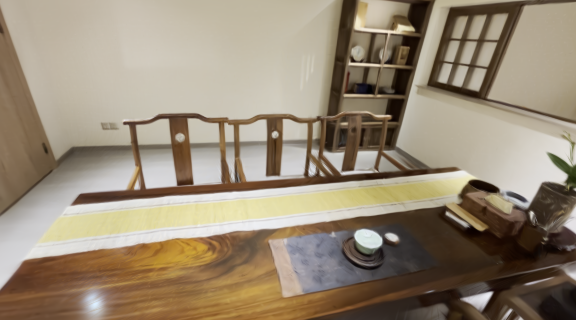} &
    \includegraphics[width=0.19\textwidth]{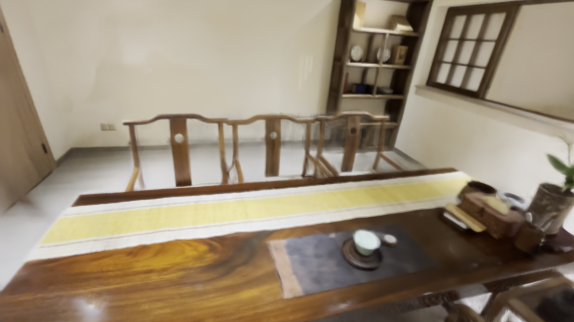} &
    \includegraphics[width=0.19\textwidth]{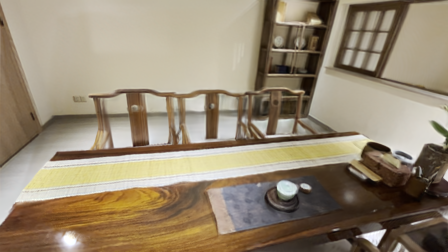} \\
    \rotatebox{90}{View 3} &
    \includegraphics[width=0.19\textwidth]{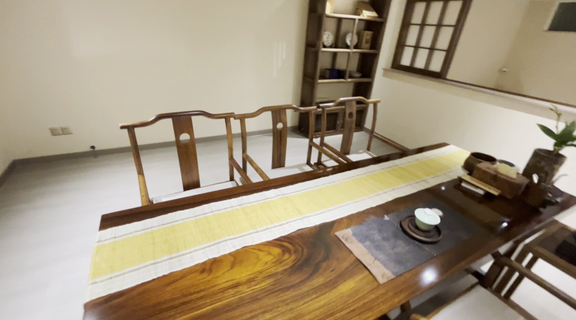} &
    \includegraphics[width=0.19\textwidth]{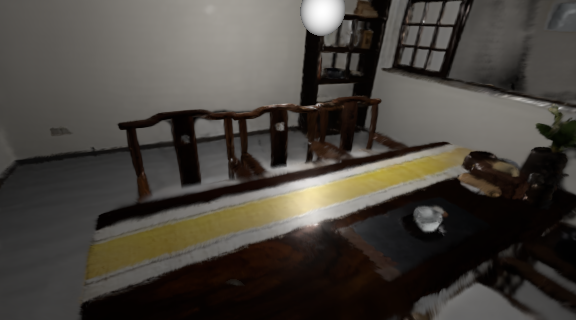} &
    \includegraphics[width=0.19\textwidth]{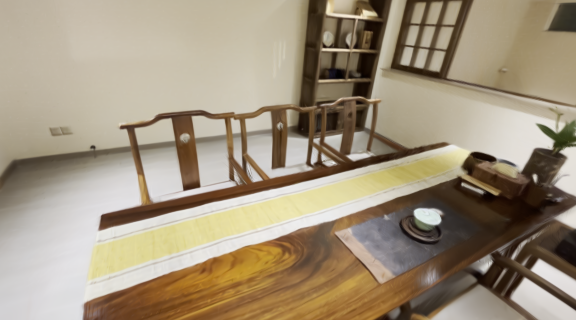} &
    \includegraphics[width=0.19\textwidth]{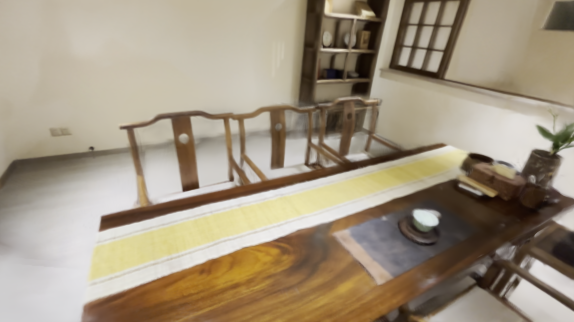} &
    \includegraphics[width=0.19\textwidth]{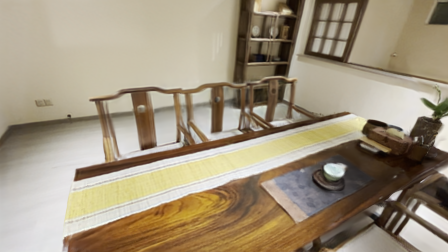} \\
\end{tabular}
  \caption{RGB reconstruction comparison on DL3DV across four views (rows) and methods (columns). Although we don't estimate scene lighting and can't produce exactly the same RGB reconstruction, our method can disentangle light from appearance and our highlights move with view change, while RGB-based methods struggle with this.}
  \label{fig:relighting_real}
\end{figure*}

When ground-truth lighting is available (as in our Infinigen scenes), this allows the renderer to reproduce view-dependent effects by construction, leading to more faithful relit RGB renderings and improved consistency across novel viewpoints. We provide an example on synthetic scene generated with Infinigen in Figure~\ref{fig:relighting_infinigen}. In this example, we knew the light configuration and used it to render the reconstructed Gaussians, which allows us to faithfully reproduce view-dependent effects such as specular highlights. In contrast, AnySplat and WorldMirror, which predict RGB appearance directly, struggle to capture these effects and tend to bake them into color. ReSplat has similar effect, although it it less visible than on other methods. Additionally, we show a similar effect on real-world scenes in Figure~\ref{fig:relighting_real}, where we don't have access to ground-truth lighting. In this case, we used a point light source and placed it in the middle of the scene. Although our model doesn't reproduce the original RGB image, we can see realistic highlight movements, while other methods struggle to capture them. This shows that our model can learn to separate view-dependent effects from albedo, producing more realistic relighting results. 

\subsection{Limitations}
\begin{wrapfigure}[7]{r}{0.4\linewidth}
\centering
\setlength{\tabcolsep}{0pt}
\renewcommand{\arraystretch}{0}
\resizebox{\linewidth}{!}{%
\begin{tabular}{@{}cc@{}}
  \includegraphics[width=0.48\linewidth]{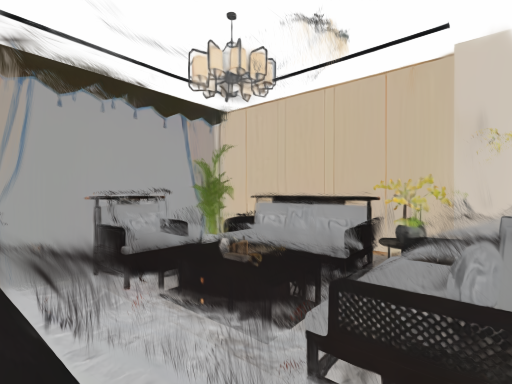} &
  \includegraphics[width=0.48\linewidth]{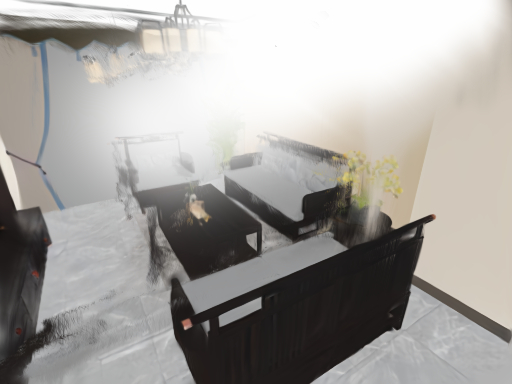} \\[4pt]
  \small View 1 & \small View 2 \\
\end{tabular}
}
\caption{Failure case example.}
\label{fig:limitation_bad_poses}
\end{wrapfigure}
Our model inherits the limitations of 2 domains. From the feed-forward scene reconstruction side, if poses are estimated incorrectly, our reconstruction will produce corrupted results, as illustrated in Figure~\ref{fig:limitation_bad_poses}. Also, performance on more views and different resolutions is limited. Additionally, as our feed-forward model is non-generative, it lacks the ability to reconstruct unseen regions of a scene with detailed geometry and texture.

\subsection{Broader Impacts}

Our feed-forward inverse rendering model significantly accelerates 3D content creation, democratizing asset generation for AR/VR while reducing the heavy energy consumption associated with traditional per-scene optimization. However, this increased accessibility also lowers the barrier for malicious applications, such as generating deceptive 3D media or unauthorized replication of intellectual property. Additionally, our model may inherit biases from its training data. Mitigating these risks requires ongoing community efforts in 3D provenance and diverse dataset curation.

\subsection{Real-world generalizability}
\begin{figure*}[p]
  \centering
  \setlength{\tabcolsep}{0pt}
  \renewcommand{\arraystretch}{0}
  \begin{tabular}{@{}>{\centering\arraybackslash}m{0.03\textwidth} *{6}{>{\centering\arraybackslash}m{0.165\textwidth}}@{}}
    & Input & Albedo & Metallic & Roughness & Normal & Depth \\[2pt]
    \rotatebox{90}{View 1} &
    \includegraphics[width=0.165\textwidth]{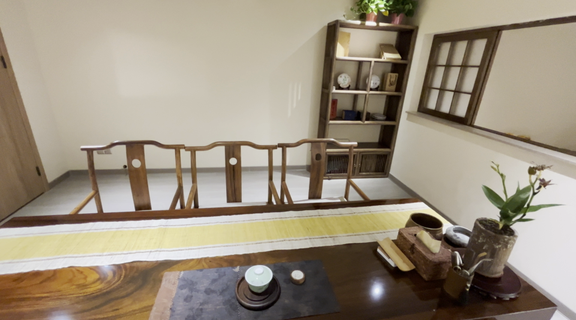} &
    \includegraphics[width=0.165\textwidth]{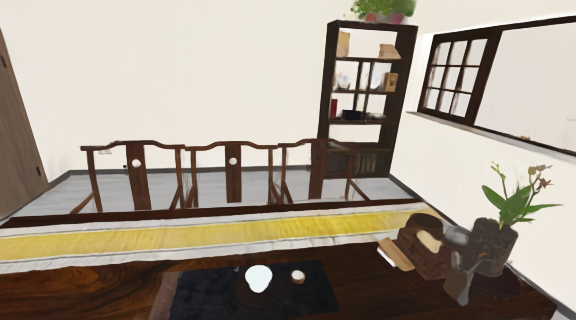} &
    \includegraphics[width=0.165\textwidth]{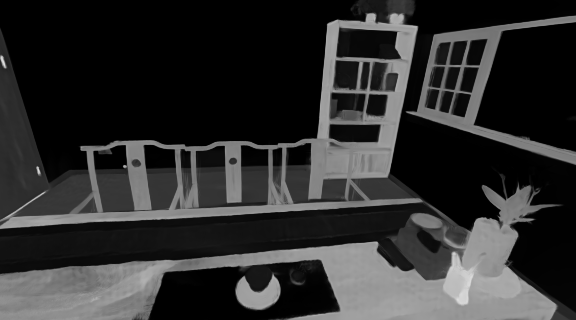} &
    \includegraphics[width=0.165\textwidth]{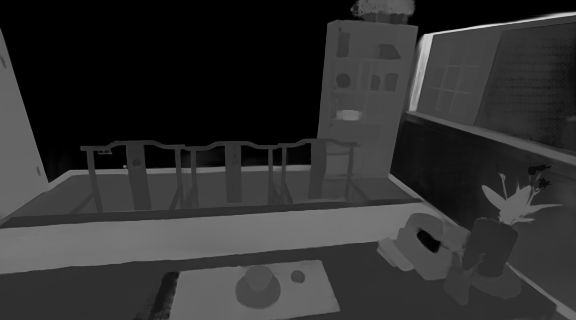} &
    \includegraphics[width=0.165\textwidth]{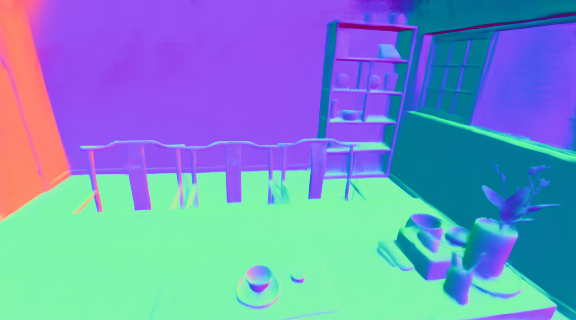} &
    \includegraphics[width=0.165\textwidth]{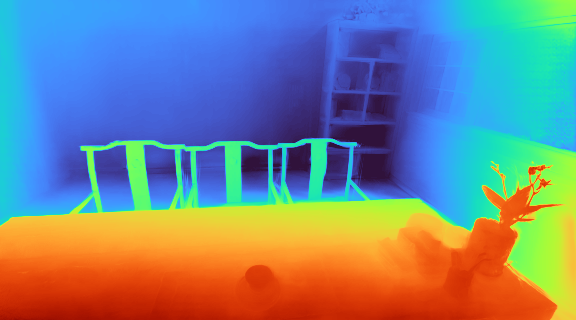} \\
\end{tabular}
  \caption{Qualitative results on three DL3DV scenes with 2 input views each. For every scene the first two rows show the predicted intrinsics (albedo, metallic, roughness, normal, depth) at the two input views. The third row shows novel-view synthesis results at an interpolated viewpoint between them.}
  \label{fig:dl3dv_nvs_example}
\end{figure*}

\begin{figure*}[p]
  \centering
  \setlength{\tabcolsep}{0pt}
  \renewcommand{\arraystretch}{0}
  \begin{tabular}{@{}>{\centering\arraybackslash}m{0.03\textwidth} *{6}{>{\centering\arraybackslash}m{0.165\textwidth}}@{}}
    & Input & Albedo & Metallic & Roughness & Normal & Depth \\[2pt]
    \rotatebox{90}{View 1} &
    \includegraphics[width=0.165\textwidth]{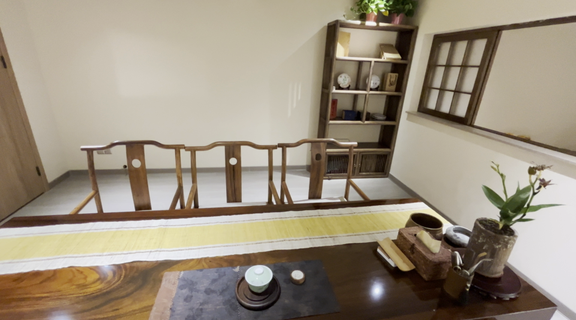} &
    \includegraphics[width=0.165\textwidth]{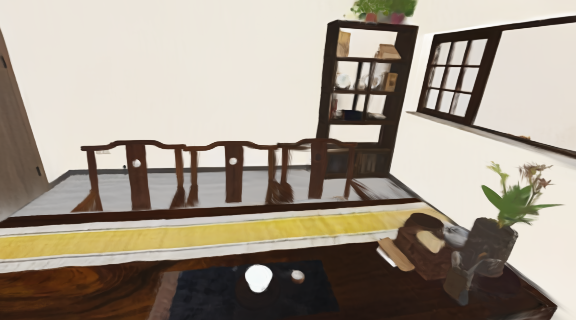} &
    \includegraphics[width=0.165\textwidth]{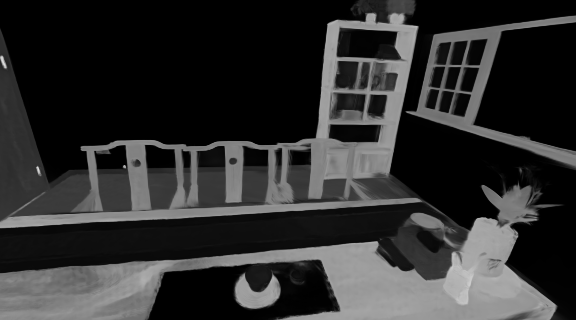} &
    \includegraphics[width=0.165\textwidth]{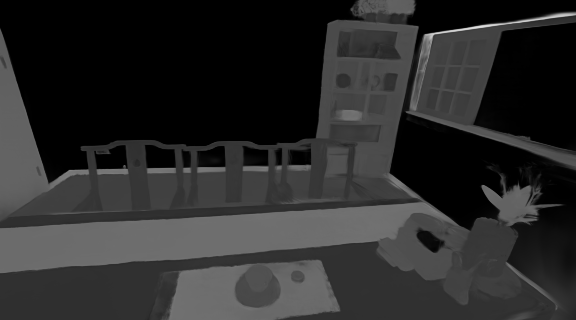} &
    \includegraphics[width=0.165\textwidth]{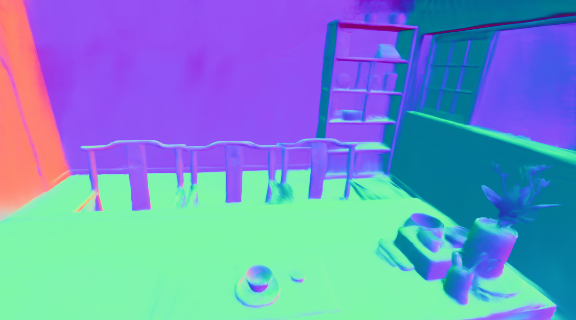} &
    \includegraphics[width=0.165\textwidth]{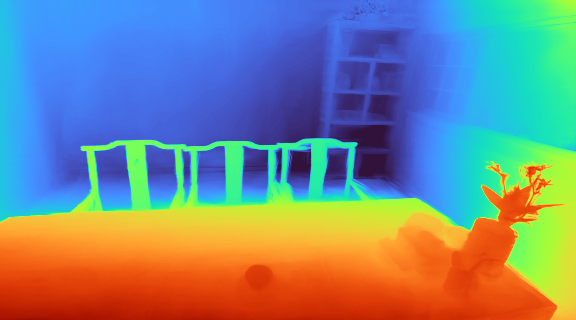} \\
\end{tabular}
  \caption{Qualitative results on three DL3DV scenes with 4 input views each. For every scene the first four rows show the predicted intrinsics (albedo, metallic, roughness, normal, depth) at the four input views. The fifth row shows novel-view synthesis results at an interpolated viewpoint between the first two input views.}
  \label{fig:dl3dv_nvs_example_views4}
\end{figure*}

We provide results of inverse rendering of our model on DL3DV dataset with 2 input views in \Cref{fig:dl3dv_nvs_example} and with 4 input views in \Cref{fig:dl3dv_nvs_example_views4}.

\begin{figure*}[t]
  \centering
  \setlength{\tabcolsep}{2pt}
  \renewcommand{\arraystretch}{0}
  \resizebox{\textwidth}{!}{%
  \begin{tabular}{@{}>{\centering\arraybackslash}m{0.025\textwidth} *{6}{>{\centering\arraybackslash}m{0.155\textwidth}}@{}}
    & \textbf{Input} & \textbf{Albedo GT} & \textbf{Ours} & \textbf{MVInverse} & \textbf{DiffRenderer} & \textbf{DNF-intrinsic} \\
    \rotatebox{90}{View 1} &
    \includegraphics[width=0.155\textwidth]{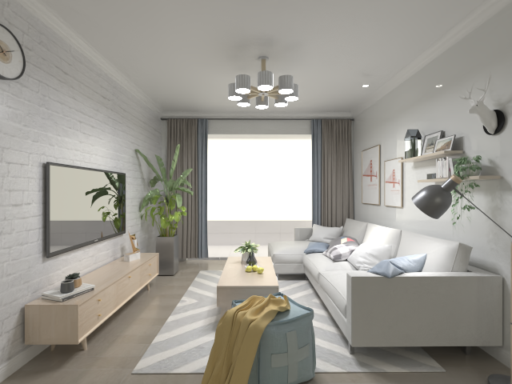} &
    \includegraphics[width=0.155\textwidth]{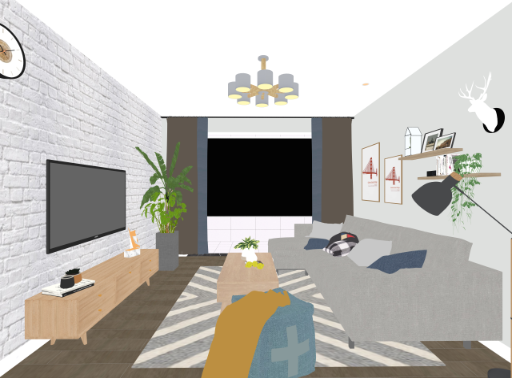} &
    \includegraphics[width=0.155\textwidth]{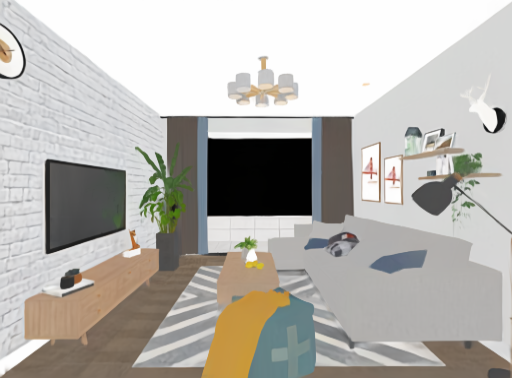} &
    \includegraphics[width=0.155\textwidth]{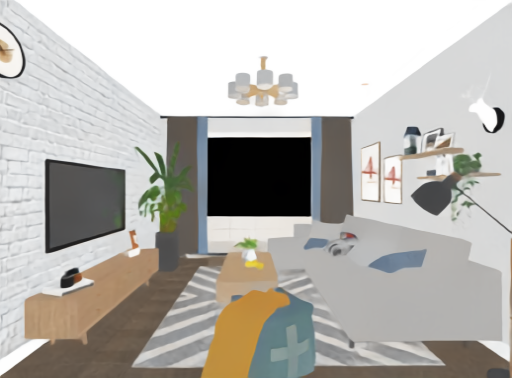} &
    \includegraphics[width=0.155\textwidth]{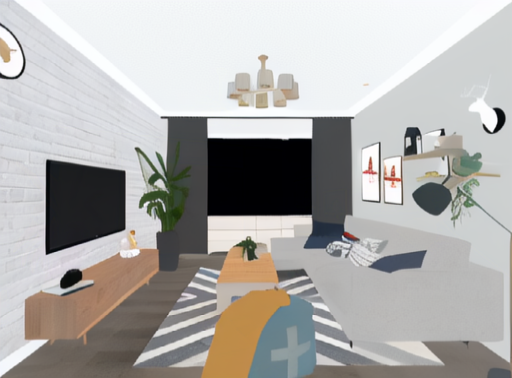} &
    \includegraphics[width=0.155\textwidth]{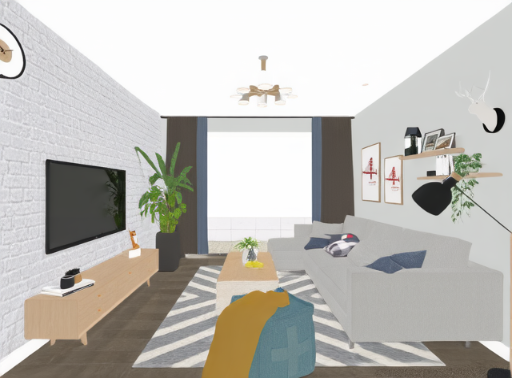} \\
    \rotatebox{90}{View 2} &
    \includegraphics[width=0.155\textwidth]{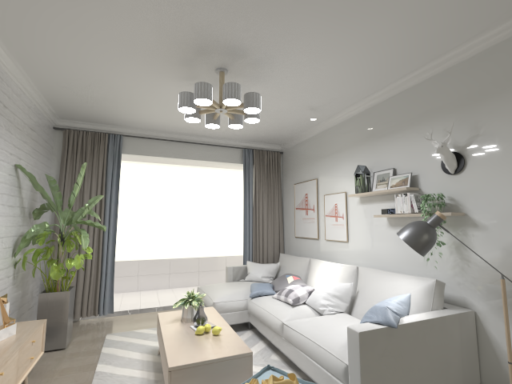} &
    \includegraphics[width=0.155\textwidth]{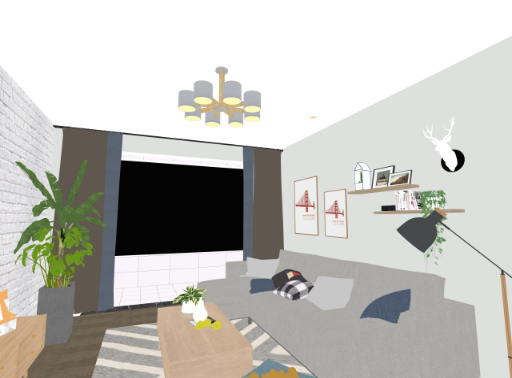} &
    \includegraphics[width=0.155\textwidth]{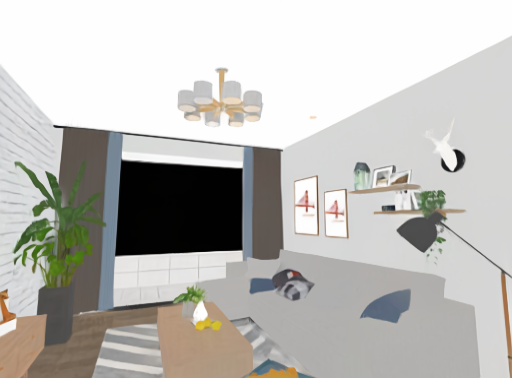} &
    \includegraphics[width=0.155\textwidth]{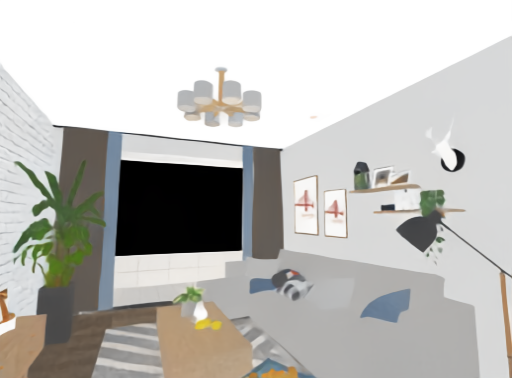} &
    \includegraphics[width=0.155\textwidth]{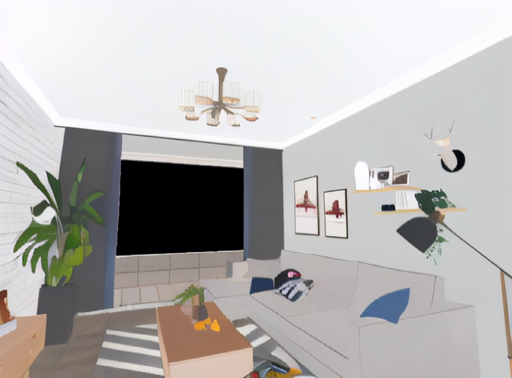} &
    \includegraphics[width=0.155\textwidth]{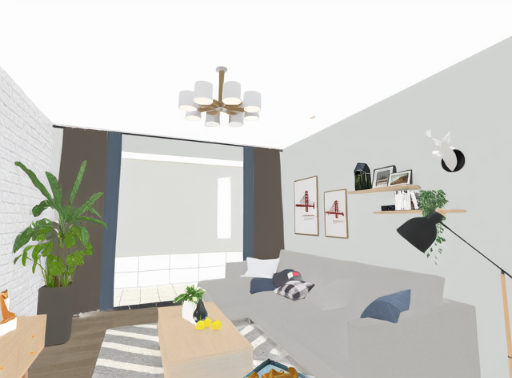} \\
\end{tabular}%
  }
  \caption{Qualitative albedo comparison on InteriorVerse. Each row shows one view of a scene. Columns: input RGB, albedo ground truth, and predicted albedo from different methods.}
  \label{fig:interiorverse_qual}
\end{figure*}

\begin{figure*}[p]
  \centering
  \setlength{\tabcolsep}{0pt}
  \renewcommand{\arraystretch}{0}
  \begin{tabular}{@{}>{\centering\arraybackslash}m{0.025\textwidth} *{6}{>{\centering\arraybackslash}m{0.16\textwidth}}@{}}
    & \textbf{Input} & \textbf{GT} & \textbf{Ours} & \textbf{MVInverse} & \textbf{DiffRenderer} & \textbf{DNF-intrinsic} \\[2pt]
    \multirow{2}{*}{\rotatebox{90}{Metallic}} &
    \includegraphics[width=0.16\textwidth]{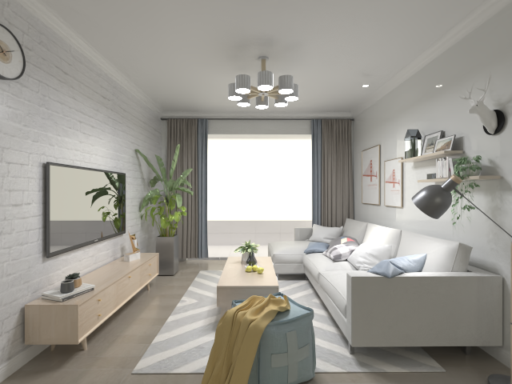} &
    \includegraphics[width=0.16\textwidth]{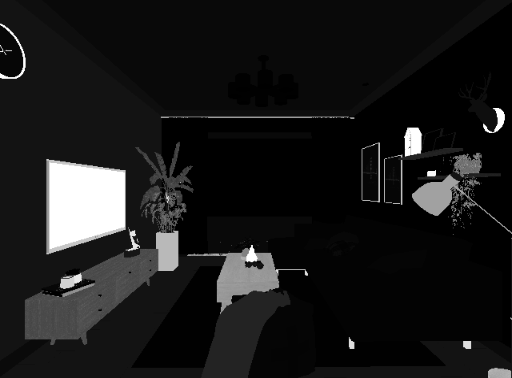} &
    \includegraphics[width=0.16\textwidth]{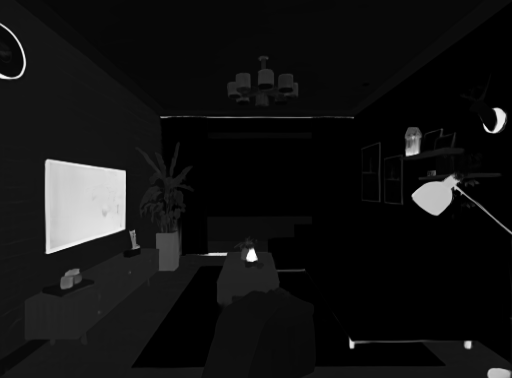} &
    \includegraphics[width=0.16\textwidth]{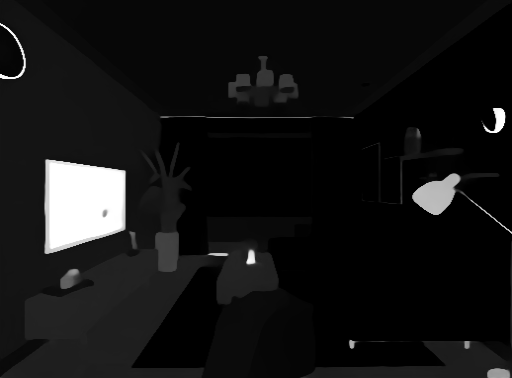} &
    \includegraphics[width=0.16\textwidth]{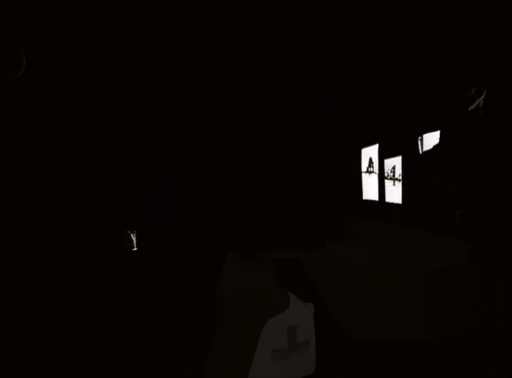} &
    \includegraphics[width=0.16\textwidth]{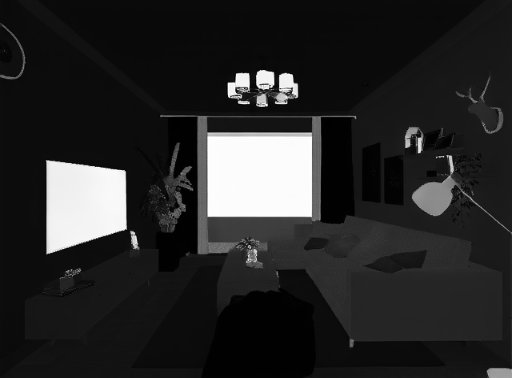} \\
    &
    \includegraphics[width=0.16\textwidth]{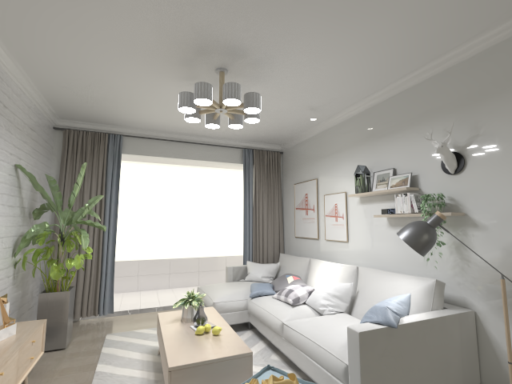} &
    \includegraphics[width=0.16\textwidth]{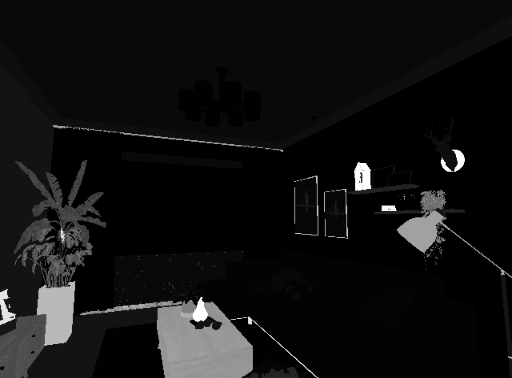} &
    \includegraphics[width=0.16\textwidth]{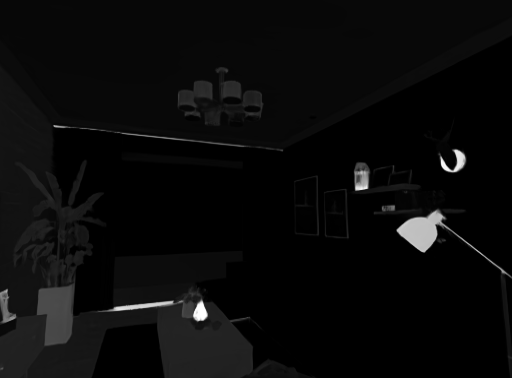} &
    \includegraphics[width=0.16\textwidth]{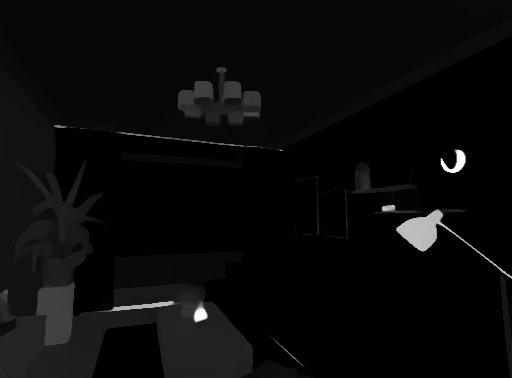} &
    \includegraphics[width=0.16\textwidth]{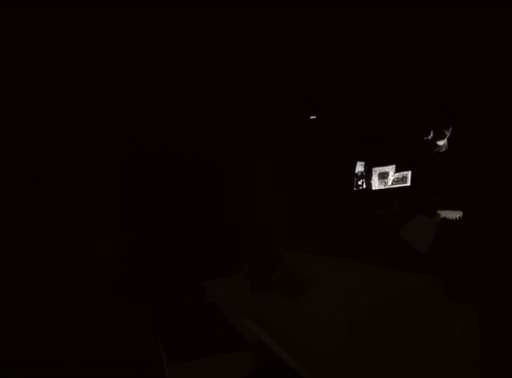} &
    \includegraphics[width=0.16\textwidth]{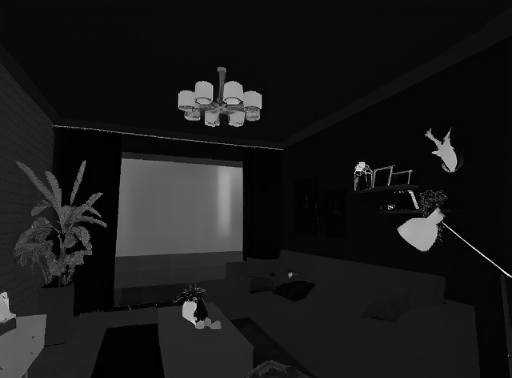} \\
\end{tabular}
  \caption{Qualitative metallic and roughness comparison on InteriorVerse for the same 3 scenes as Figure~\ref{fig:interiorverse_qual}. For each scene we show 2 metallic rows followed by 2 roughness rows (one row per view). Columns: input RGB, ground truth, and predictions from each method.}
  \label{fig:interiorverse_metallic_roughness}
\end{figure*}

\begin{figure*}[p]
  \centering
  \setlength{\tabcolsep}{0pt}
  \renewcommand{\arraystretch}{0}
  \resizebox{0.9\textwidth}{!}{%
  \begin{tabular}{@{}>{\centering\arraybackslash}m{0.025\textwidth}@{\hspace{2pt}}*{3}{>{\centering\arraybackslash}m{0.145\textwidth}}@{\hspace{8pt}}*{3}{>{\centering\arraybackslash}m{0.145\textwidth}}@{}}
    & \multicolumn{3}{c}{\footnotesize \textbf{Ours}} & \multicolumn{3}{c}{\footnotesize \textbf{MVInverse}} \\[6pt]
    & {\footnotesize View 0} & {\footnotesize Warp 1$\to$0} & {\footnotesize Error} & {\footnotesize View 0} & {\footnotesize Warp 1$\to$0} & {\footnotesize Error} \\[6pt]
    \rotatebox{90}{\footnotesize Albedo} &
    \includegraphics[width=0.145\textwidth]{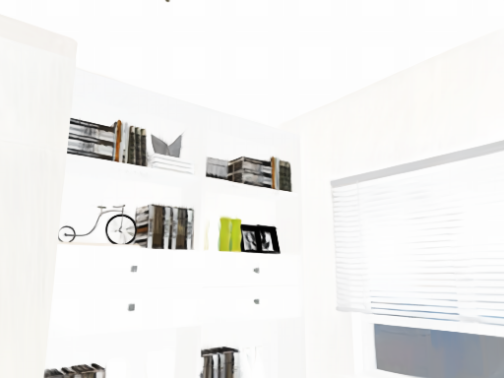} &
    \includegraphics[width=0.145\textwidth]{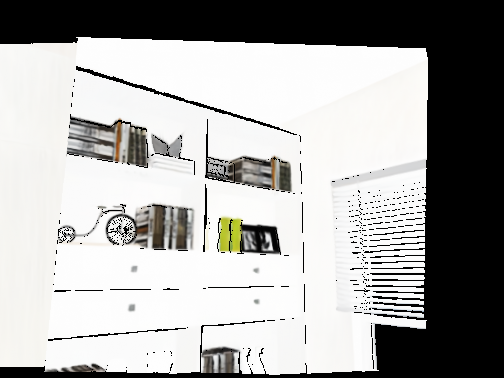} &
    \includegraphics[width=0.145\textwidth]{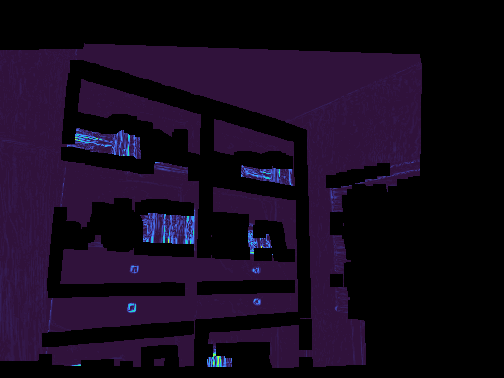} &
    \includegraphics[width=0.145\textwidth]{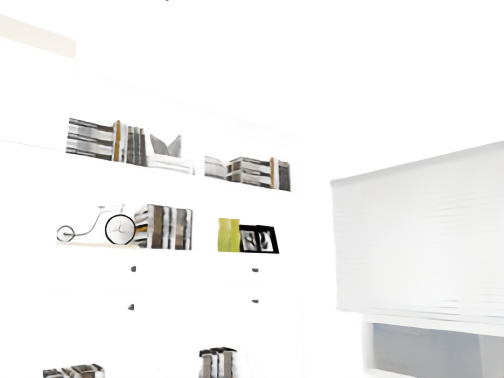} &
    \includegraphics[width=0.145\textwidth]{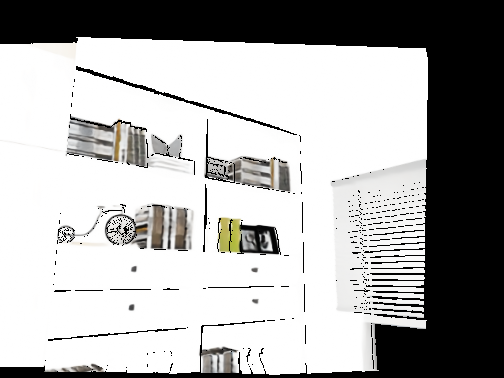} &
    \includegraphics[width=0.145\textwidth]{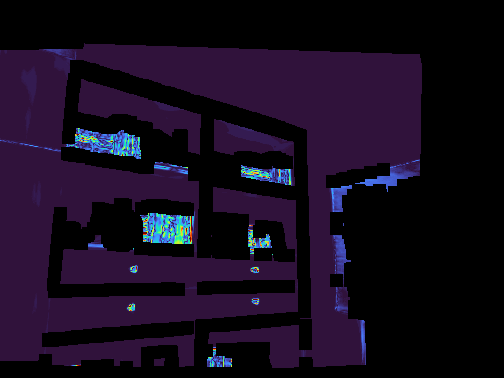} \\
\end{tabular}%
  }
  \caption{Multi-view consistency on additional examples from Structured3D for albedo, metallic and roughness. For each method, the figure shows the prediction at view 0, the prediction from view 1 warped into view 0 using ground-truth depth and pose, and the per-pixel error between them.}
  \label{fig:consistency_structured3d_compressed_all}
\end{figure*}

\newpage



\end{document}